%% file: main.tex
\pdfoutput=1
\documentclass[11pt]{article}

\usepackage[final]{acl}

\usepackage{times}
\usepackage{latexsym}
\usepackage{graphicx}

\usepackage[T1]{fontenc}
\usepackage[utf8]{inputenc}

\usepackage{microtype}

\usepackage{inconsolata}

\usepackage{graphicx}
\usepackage{booktabs}
\usepackage{adjustbox}
\usepackage{multirow}
\usepackage{amsmath}
\usepackage{amssymb}

\usepackage{cleveref}
\usepackage{algorithm}
\usepackage{algpseudocode}
\usepackage{xargs}
\usepackage{makecell}
\usepackage{enumitem}
\usepackage{longtable}
\usepackage{tcolorbox}
\usepackage{multicol}
\usepackage{setspace}
\usepackage{makecell}
\usepackage{fontawesome}
\usepackage{pgfplots}
\usepackage{pgfplotstable}
\usepackage{caption}
\usepackage{subcaption}

\pgfplotsset{compat=1.17}

\newcommand{\renact}{ReNAct}
\newcommand{\noteswriting}{\textit{NotesWriting}}
\newcommand{\react}{ReAct}

\expandafter\def\expandafter\normalsize\expandafter{%
    \normalsize%
    \setlength\abovedisplayskip{3pt}%
    \setlength\belowdisplayskip{3pt}%
    \setlength\abovedisplayshortskip{3pt}%
    \setlength\belowdisplayshortskip{3pt}%
}

\title{Augmenting LLM Reasoning with Dynamic Notes Writing for Complex QA}

\author{
Rishabh Maheshwary\textsuperscript{\S} \hspace{0.5em}
Masoud Hashemi\textsuperscript{\S} \hspace{0.5em}
Khyati Mahajan\textsuperscript{\S} \hspace{0.5em}
\\
\textbf{Shiva Krishna Reddy Malay}\textsuperscript{\S} \hspace{0.5em}
\textbf{Sai Rajeswar}\textsuperscript{\dag} \hspace{0.5em}
\textbf{Sathwik Tejaswi Madhusudhan}\textsuperscript{\S} \hspace{0.5em}
\\
\textbf{Spandana Gella}\textsuperscript{\dag} \hspace{0.5em}
\textbf{Vikas Yadav}\textsuperscript{\S} \\
\textsuperscript{\S} ServiceNow \\
\textsuperscript{\dag} ServiceNow Research \\
 \texttt{\{rishabh.maheshwary, vikas.yadav\}} @servicenow.com$^{\S}$
}

\begin{document}
\maketitle
\begin{abstract}

Iterative RAG for multi-hop question answering faces challenges with lengthy contexts and the buildup of irrelevant information. This hinders a model's capacity to process and reason over retrieved content and limits performance. While recent methods focus on compressing retrieved information, they are either restricted to single-round RAG, require finetuning or lack scalability in iterative RAG. To address these challenges, we propose \noteswriting, a method that generates concise and relevant notes from retrieved documents at each step, thereby reducing noise and retaining only essential information. This indirectly increases the \emph{effective context length} of Large Language Models (LLMs), enabling them to reason and plan more effectively while processing larger volumes of input text. \noteswriting~is framework agnostic and can be integrated with different iterative RAG methods. We demonstrate its effectiveness with three iterative RAG methods, across two models and four evaluation datasets. 
\noteswriting~yields an average improvement of 15.6 percentage points overall, with minimal increase in output tokens. 
% Additionally, we advocate leveraging ReAct as a powerful framework to formalize and standardize the iterative RAG flow.
% stemming from retrieved information accumulated across steps, leading to redundancy and inefficient queries at later reasoning steps which degrade subsequent retrieval, and thus the overall response quality. We introduce \renact~(Reason, iterative \noteswriting, and Action) which enhances long context based multi-hop QA reasoning by iteratively retrieving and accumulating only the most relevant information at each reasoning step. \noteswriting~dynamically maintains a concise list of the relevant information which enables more effective query writing at each reasoning step allowing focus on missing information rather than reprocessing previously retrieved long content. By writing concise notes and guiding query formulation, \renact~ significantly improves both effectiveness and efficiency in multi-hop reasoning. Our approach achieves >20\% absolute F1 score gains on long-context benchmarks such as FanOutQA and FRAMES, while reducing the number of reasoning steps by 56\% on average compared to the \reactbm~baseline.

\end{abstract}

\input{latex/1Introduction}

\input{latex/2_3_Backgroundandrelatedworks}

\input{latex/4Results}

\input{latex/5_additional_results}

\input{latex/6_Conclusion}

\bibliography{custom}

\appendix

\input{latex/7_Appendix}

\end{document}

%% file: latex/1Introduction.tex
\section{Introduction} \label{sec:intro}
% Compared to Retrieve, Reason and plan paper:
% Ours is a complementary method that can be used with any baseline. Our method is simple and can be coupled with any of thre iterative retrieval baseline. Their it use multiple modules Reasoner, Retriever, Summarizer, and Generator
% Their the summarization module is focused on compressing the information, however, ours in addition to compressing is also focused on finding or extracting missing information.
% Max iteration is set to only 3 in their method.
% There setup uses wikipedia dump with presegmented passages. However, our method is able to ingest Wikipedia web articles (in Readme format), thus enabling it to look through more information.

The retrieval augmented generation (RAG) paradigm has advanced open domain question answering~\cite{zhang2022survey, kamalloo2023evaluating} by incorporating external knowledge~\cite{lewis2020retrieval, guu2020retrieval, borgeaud2022improving, shi2023replug, izacard2023atlas}, enabling Large Language Models (LLMs)~\cite{hurst2024gpt,dubey2024llama} to refresh outdated parametric knowledge~\cite{dhingra2022time, kasai2023realtime, vu2023freshllms} and mitigate hallucinations~\cite{ji2023survey, zhang2023siren}.

However, for tasks like multi hop question answering~\cite{yang-etal-2018-hotpotqa, zhu-etal-2024-fanoutqa, krishna2024fact} which requires reasoning over multiple documents, a single-round RAG based solely on the initial question often falls short, as it fails to capture all the necessary information. To overcome this, iterative RAG methods such as IRCoT~\cite{trivedi2022interleaving}, FLARE~\cite{jiang2023active}, and ReAcT~\cite{yao2023react} interleave retrieval and reasoning over multiple steps, progressively accumulating the evidence needed to answer complex queries.

Nevertheless, retrieved information can be noisy, and prior work has shown that excessive noise in the retrieved context can significantly degrade RAG performance~\cite{petroni2020context, shi2023large, zhang2024raft, leng2024long, wu2024easily}. This challenge is amplified in iterative retrieval settings, where new information must be retrieved at each reasoning step. Therefore, simply concatenating all retrieved documents at each step leads to several problems:
\begin{itemize}[noitemsep, nolistsep, leftmargin=*]
    \item \textbf{Context Overload:} Exceeding the LLM's context window limit~\cite{krishna2024fact}.
    \item \textbf{Computational Cost \& Scalabiity:} Increasing processing time and resources ~\cite{yue2024inference}.
    \item \textbf{Distraction:} Including irrelevant or redundant information that hinders the LLM's reasoning and planning ability \cite{yu2023chain, chen2024can, xie2024revealing, aghzal2025survey}.
    \item \textbf{Readability:} Excessively long reasoning traces created from multiple documents pose challenges for users to interpret precise reasoning. Redundant information can further affect readability.
\end{itemize}

To address these issues, we propose a simple yet effective and scalable method called \noteswriting. At each retrieval step, \noteswriting\ produces concise notes based upon the retrieved documents and the sub-question, thus providing only the essential information required at each step. This increases the \emph{effective context length} as it does not overload the LLM context with irrelevant information, which helps the LLMs plan \& reason better. Furthermore, \noteswriting~is generic and can be coupled with any iterative RAG framework.
\noindent
While recent methods~\cite{edge2024local, xu2023recomp, kim2024sure} have explored summarizing retrieved content, they are often limited to single-round RAG or require synthetic data for fine-tuning. \citet{jiang2025retrievesummarizeplanadvancing} extend this idea by summarizing retrieved documents at each step; however, this approach lacks scalability as it is still limited to three iterations at maximum and depends on multiple modules leading to multiple LLM calls in each iteration. 

\noindent
This work makes the following key contributions:
\begin{itemize} [nolistsep, noitemsep, leftmargin=*]
    \item We propose \noteswriting\ to improve effective context length of iterative RAG. This reduces context overload, the number of reasoning steps, and redundancy.
    
    % \item This increases the \emph{effective context length} as it does not overload the LLM context with noise or irrelevant information which helps the LLMs plan \& reason better. Furthermore, our method is generic and can be coupled with any iterative RAG framework.

    \item \noteswriting\ is plug-and-play, and can be coupled with any iterative RAG framework, benefiting planning and reasoning abilities by reducing tokens in context (thus indirectly increasing the effective context length per step) and reducing retention of irrelevant information.

    \item Our experiments across \emph{three} iterative RAG baselines (IRCoT, FLARE \& \react), \emph{four} multi-hop QA datasets and \emph{two} LLMs demonstrates that \noteswriting\ achieves 15.6 percentage points improvement by increasing the volume of ingested text with minimal increase in output tokens. %We further show that with ReAcT being the best iterative framework, adding \noteswriting\ helps in stepwise decision making and the reasoning chains are more efficient, concise and clearer.

    \item \noteswriting\ with \react\ (\renact) achieves the highest performance, enabling better planning by guiding the model to generate more accurate search queries and retrieving correct documents as demonstrated in Section~\ref{sec:analysis}.
    \end{itemize}

%% file: latex/2_3_Backgroundandrelatedworks.tex
\begin{figure*}[th!]
\centering
 \includegraphics[width=\linewidth]{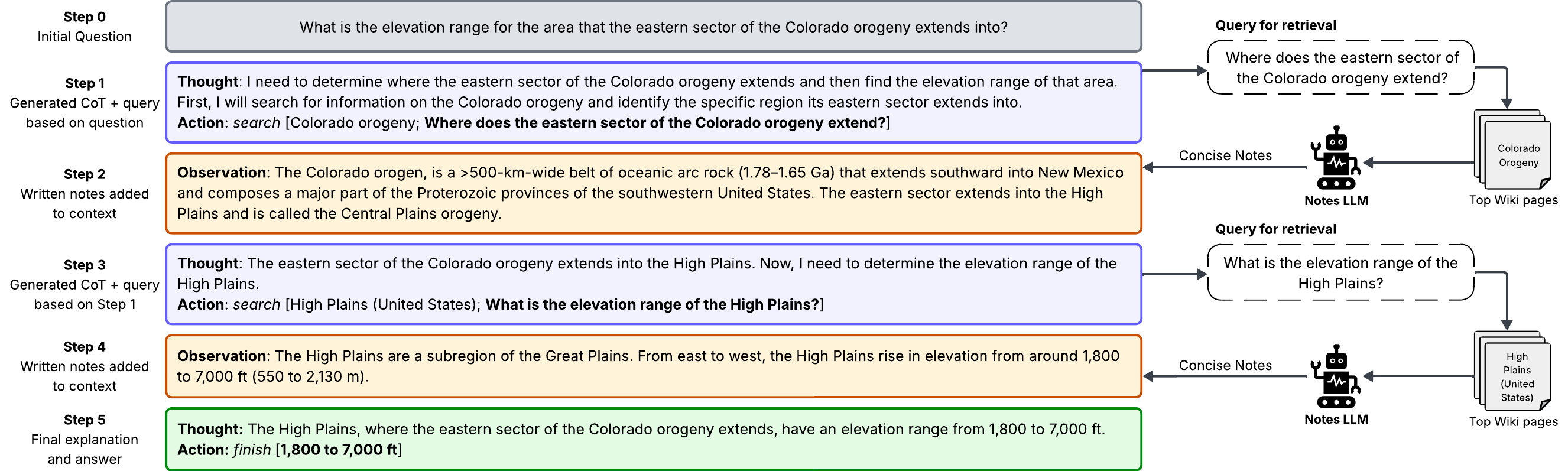}
 \caption{Overview of \noteswriting~within an iterative RAG framework.}
 \label{fig:overview}
\end{figure*}

\section{Background and Related Work} \label{sec:related-work}

\paragraph{Single-Step vs. Iterative RAG.}
Traditional RAG often operates in a single step: retrieve relevant documents based on the initial query, then generate the final response conditioned on both the query and the retrieved context. While effective for simpler questions, this retrieve-then-read approach struggles for multi-hop QA, where the information needed evolves throughout the reasoning process. Iterative RAG addresses this limitation by interleaving retrieval and generation. The model can issue multiple queries, gather information incrementally, and refine its reasoning path based on newly retrieved evidence. This dynamic interaction between the LLM and the retriever is better suited for complex, multi-step reasoning.

\paragraph{Formulation of Iterative RAG.}

Let $\mathbf{x}$ be the user input question, and $\mathcal{D} = \{d_i\}_{i=1}^{|\mathcal{D}|}$ represent the external knowledge corpus (e.g., Wikipedia). An iterative RAG process aims to generate a sequence of reasoning steps or partial outputs $\mathbf{s} = [\mathbf{s}_1, \mathbf{s}_2, \dots, \mathbf{s}_n]$. We denote the language model as $\text{LM}(\cdot)$ and the retrieval function, which returns the top-$k$ documents for a query $\mathbf{q}$, as $\text{ret}(\mathbf{q})$.

At each step $t \geq 1$, the typical process involves:
\begin{enumerate} [nolistsep, noitemsep, leftmargin=*]
    \item \textbf{Query Formulation:} A query $\mathbf{q}_t$ is generated based on the initial input $\mathbf{x}$ and the preceding steps $\mathbf{s}_{<t} = [\mathbf{s}_1, \dots, \mathbf{s}_{t-1}]$. This is governed by a query formulation function $\mathcal{Q}(\cdot)$:
    % \vspace{-1mm}
    \begin{equation}
    \mathbf{q}_t = \mathcal{Q}(\mathbf{x}, \mathbf{s}_{<t})
    \end{equation}
    For the first step, $\mathbf{s}_{<1} = \emptyset$, and often $\mathbf{q}_1 = \mathbf{x}$.
    \item \textbf{Retrieval:} The retriever fetches the top-$k$ relevant documents: $\mathcal{D}_{\mathbf{q}_t} = \text{ret}(\mathbf{q}_t)$.
    \item \textbf{Generation:} The LM generates the next reasoning step $\mathbf{s}_t$ using the original input, previous steps, and the newly retrieved documents:
    \begin{equation}
    \mathbf{s}_t = \text{LM}([\mathcal{D}_{\mathbf{q}_t}, \mathbf{x}, \mathbf{s}_{<t}])
    \end{equation}
\end{enumerate}    
% \vspace{-1mm}
This process continues until a final answer is generated, or a maximum number of steps is reached.

\paragraph{Advances in Iterative RAG.}

Several approaches have explored different strategies within this iterative framework:
IRCoT (Interleaving Retrieval and Chain-of-Thought) \cite{trivedi2022interleaving} directly uses the last generated sentence of the Chain-of-Thought (CoT) \cite{wei2022chain} reasoning as the query for the next retrieval step. This tightly couples reasoning progress with information gathering. Formally, its query function is:
\begin{equation}
\mathbf{q}_t = \mathcal{Q}_{\text{IRCoT}}(\mathbf{x}, \mathbf{s}_{<t}) = \mathbf{s}_{t-1}
\end{equation}
IRCoT demonstrates improved performance over single-step RAG across various QA datasets, showing robustness even with smaller models. \\ 
FLARE (Forward-Looking Active Retrieval Augmented Generation) \cite{jiang2023active} builds on similar ideas but retrieves proactively. It anticipates future content needs by checking the confidence of generated tokens. Retrieval is triggered only when the model generates low-confidence tokens (often indicating a need for information), using the preceding sentence as the query. If confidence is high, retrieval is skipped. The query function is conditional:
\begin{equation}
\mathbf{q}_t =
\begin{cases}
\mathbf{s}_{t-1}, & \text{if } \text{conf}(\mathbf{s}_{t-1}) > \theta \\
\emptyset, & \text{otherwise}
\end{cases}
\end{equation}
While effective in long-form generation, its reliance on predicting future needs might face challenges if the context shifts unexpectedly.\\
ReAct (Reason and Action)~\cite{yao2023react} explicitly separates reasoning (\texttt{Thought}) from information gathering (\texttt{Action}), where the action often involves generating a specific search query.
\begin{equation} \label{eq:generation}
\mathbf{q}_t = \mathcal{Q}_{\text{ReAct}}(\mathbf{x}, \mathbf{s}_{<t}) = \texttt{Action}_t
\end{equation}

% \paragraph{Long context LLMs and RAG.} Long context LLMs have emerged as a possible solution to the need for managing multiple streams of information needed for multi-hop QA. 

\paragraph{Context Management in Iterative RAG.} Despite advancements in iterative RAG, a core challenge persists: managing the retrieved context effectively across iterations. Even with long context LLMs, studies have found that complex tasks which require compositional reasoning like multi-hop QA are solved better with retrieval \cite{xu2023retrieval,lee2024can}. However, long context LLMs have been shown to face issues in handling information within the long context (needle-in-the-haystack issues) \cite{kamradt2023needle,hsieh2024ruler} which limit performance even when combined with RAG \cite{jiang2024longrag}.
%Simply concatenating all retrieved documents ($\mathcal{D}_{\mathbf{q}_t}$) at each step lead to several problems:
% \begin{itemize}[noitemsep,nolistsep]
%     \item \textbf{Context Overload:} Exceeding the LLM's context window limit.
%     \item \textbf{Computational Cost:} Increasing processing time and resources.
%     \item \textbf{Distraction:} Including irrelevant or redundant information that hinders the LLM's reasoning \cite{yu2023chain}.
%     \item \textbf{Readability:} Excessively long reasoning traces created from multiple documents pose challenges for users to interpret the precise reasoning. Multiple documents could also contain redundant information further affecting readability of human users.
% \end{itemize}
% \noindent
Thus, addressing context management requires mechanisms to condense, filter, or summarize the retrieved information. Several such approaches have been explored in recent research. RECOMP \cite{xu2023recomp} compresses retrieved documents using extractive or abstractive summarization before passing them to the main LLM in a single-turn RAG setting. This helps with query-relevant compression, but does not directly handle iterative context accumulation. Chain-of-Note (CON) \cite{yu2023chain} generates sequential notes during training to assess retrieved document relevance and reliability. This improves robustness against noise, but lacks explicit planning or iterative refinement at inference time. PlanRAG \cite{lee2024planrag} proposes a two-stage approach, generating a decision plan and then executing retrieval operations by adding Plan and Re-plan steps to ReAct. SmartRAG \cite{gao2024smartrag} similarly includes a policy network which decides whether to retrieve, and a retriever, which are jointly optimized to reduce retrieval while improving performance. However, retrieved documents while more relevant can still accumulate in context, affecting performance. 
Self-RAG \cite{asai2023self} uses reflection tokens to self-reflect on the retrieved documents as a means to reduce the number of documents included in the context. Self-reflection is achieved by fine-tuning the model, which improves factual accuracy and citation integrity when benchmarked against existing models such as ChatGPT and Llama2-Chat. However, the requirement of fine-tuning could be costly, and require updates over time.

% GraphRAG \cite{edge2024local} uses an LLM to create a knowledge graph based on the private retrieval document dataset which is then used alongside graph machine learning to perform prompt augmentation at query time. However, this approach requires retrieval documents to be known beforehand, and can lead to retrieval of outdated information over time. Moreover, it also requires graph-specific embeddings and query languages, adding complexity distinct from text-based iteration.
% AutoRAG \cite{kim2024autorag} seeks to address the variability in performance of RAG modules across different datasets by autonomously identifying the most suitable RAG configurations for a given context. However, its reliance on dataset characteristics may pose limitations in less structured environments.
\noindent
More recently, modular approaches have been suggested for iterative retrieval and summarization. Infogent \cite{reddy2024infogent} proposes two modules, an Aggregator whose textual feedback guides further retrieval from a Navigator-Extractor. The Extractor extracts
readable relevant content from the Navigator's web-based API access and forwards it to the Aggregator for evaluation. However, context management remains an important issue. ReSP (Retrieve, Summarize, Plan) \cite{jiang2025retrievesummarizeplanadvancing} uses query-focused summarization in multi-hop QA, maintaining global and local evidence summaries across iterations to prevent context overload. It involves multiple LLM calls per iteration focusing on planning for the next step with sub-questions, summarizing retrieved documents, generating the next sub-questions, and judging for whether there is sufficient information to answer the question. While specialized modules for each stage could boost performance further, this approach faces several drawbacks - like the possibility of cascading failures if any module fails during an iteration, and multiple LLM calls which can further increase latency and information repetition with local and global evidence.

 % The challenge remains, however, in balancing the complexity of these systems with the need for efficient and effective retrieval mechanisms in the context of retrieval operations in highly complex queries where multiple relevant passages are required.

Our proposed \noteswriting~method overcomes the aforementioned challenges by focusing on flexibly generating concise and relevant notes from retrieved documents at each iterative step. This addresses the critical need for noise reduction and context length enhancement, thereby allowing LLMs to reason and plan more effectively in complex multi-hop scenarios.

\section{Method}

To address the challenges of context overload and information noise in iterative RAG, particularly for multi-hop QA, we introduce \textbf{\noteswriting}, a method for generating concise, query-relevant notes from retrieved documents at each step. Instead of feeding raw retrieved documents to the main LM, \noteswriting~first processes them to extract key information, thereby reducing context length and filtering irrelevant content.

\subsection{\noteswriting: Iterative Note Extraction}

The core idea is to use a dedicated, smaller language model ($\text{LM}_{\text{notes}}$) to act as a note-taker. At each iteration $t$, after retrieving the top-$k$ documents $\mathcal{D}_{\mathbf{q}_t} = \{d_1, d_2, \dots, d_k\}$ based on the query $\mathbf{q}_t$, \noteswriting\ performs the following:
\begin{enumerate} [nolistsep, noitemsep, leftmargin=*]
    \item \textbf{Note Extraction:} For each retrieved document $d_i$, $\text{LM}_{\text{notes}}$ is prompted (using prompt $\mathcal{P}_{\text{notes}}$, see \Cref{fig:notes_writing_prompt}) to extract concise notes $r_i$ relevant to the current query $\mathbf{q}_t$:
    \begin{equation}
    r_i = \text{LM}_{\text{notes}}(\mathbf{q}_t, d_i)
    \end{equation}
    \item \textbf{Note Aggregation:} The extracted notes from all $k$ documents are aggregated as $\mathbf{O}_t$:
    \begin{equation}
    \mathbf{O}_t = \bigcup_{i=1}^{k} r_i
    \end{equation}
    % \item \textbf{Conditioned Generation:} The main language model $\text{LM}$ then generates the next reasoning step $\mathbf{s}_t$ conditioned on the aggregated notes $\mathbf{O}_t$, the original input $\mathbf{x}$, and the previous steps $\mathbf{s}_{<t}$:
    % \begin{equation}
    % \mathbf{s}_t = \text{LM}([\mathbf{O}_t, \mathbf{x}, \mathbf{s}_{<t}])
    % \end{equation}
\end{enumerate}
This process replaces the direct feeding of potentially long and noisy documents $\mathcal{D}_{\mathbf{q}_t}$ with the much shorter and focused notes $\mathbf{O}_t$.

% \subsection{Combining ReAct with \noteswriting}
\subsection{\renact: \react\ with \noteswriting}
While \noteswriting~is a generic module that can be integrated with different iterative RAG methods, results in \Cref{sec:results} demonstrates that it works best with the \react\ framework~\cite{yao2023react}. Therefore, we propose leveraging the ReAct framework as a suitable base for our approach. \react's structure explicitly separates reasoning (\texttt{Thought}) from information gathering (\texttt{Action}), where the action often involves generating a specific search query. This explicit query generation aligns naturally with the goal of targeted retrieval followed by focused note-taking.\\
We combine \react\ with \noteswriting\ with the process at step $t$ as follows:
\begin{enumerate} [noitemsep, nolistsep, leftmargin=*]
    \item LM generates \texttt{Thought} step outlining reasoning, along with an \texttt{Action} step, typically containing a search query. This query becomes $\mathbf{q}_t$:
    \begin{equation}
    \begin{split}
    \mathbf{q}_t = & \mathcal{Q}_{\text{ReAct}}(\mathbf{x}, \mathbf{s}_{<t}) = \\
    &\text{SearchQueryFrom}(\texttt{Action}_t)
    \end{split}
    \end{equation}
    Retrieval is performed using $\mathbf{q}_t$ to get $\mathcal{D}_{\mathbf{q}_t}$.
    \item \noteswriting~ processes $\mathcal{D}_{\mathbf{q}_t}$ using $\text{LM}_{\text{notes}}$ to generate aggregated notes, presented as the observation $\mathbf{O}_t$.
    \item The main LM receives $\mathbf{O}_t$ and uses it along with $\mathbf{x}$ and $\mathbf{s}_{<t}$ to generate the next \texttt{Thought} and \texttt{Action} pair:
    \begin{equation}
    \begin{split}
    &\mathbf{s}_t (\text{next Thought+Action}) = \\
    & \qquad \text{LM}([\text{Observation: } \mathbf{O}_t, \mathbf{x}, \mathbf{s}_{<t}])
    \end{split}
    \end{equation}
\end{enumerate}
Iterations continue until the model generates a final answer within $\mathbf{s}_t$ or reaches a maximum number of iterations $T$, after which a final answer is synthesized based on the full history $\mathbf{s}$ and the collected notes $\{\mathbf{O}_t\}_{t=1}^{T}$. This approach (illustrated in \Cref{fig:overview}) aims to combine the structured reasoning of ReAct with the context management benefits of \noteswriting, leading to a more robust and efficient iterative RAG system for complex QA.

% \subsection{\noteswriting~with more Iterative RAG methods}
\subsection{\noteswriting: A Plug-and-Play Module for Iterative RAG}

\noteswriting~is designed as a complementary module that can be integrated into various iterative RAG frameworks. It modifies the generation step (Eq. \ref{eq:generation}) while keeping the specific query formulation $\mathcal{Q}$ of the base framework. We demonstrate this integration with two SOTA iterative RAG frameworks: IRCoT and FLARE.
\\\noindent
\textbf{IRCoT:} Query remains the last generated sentence ($\mathbf{q}_t = \mathbf{s}_{t-1}$). Generation step becomes:
    \begin{equation}
    \begin{split}
     \mathbf{s}_t = & \text{LM}([\mathbf{O}_t, \mathbf{x}, \mathbf{s}_{<t}]) \text{ where}\\
     \mathbf{O}_t &\text{ is derived from } \mathcal{D}_{\mathbf{q}_t} = \text{ret}(\mathbf{s}_{t-1})
    \end{split}
    \end{equation}
\textbf{FLARE:} Query formulation remains conditional based on confidence $\theta$ ($\mathbf{q}_t = \mathcal{Q}_{\text{FLARE}}(\mathbf{x}, \mathbf{s}_{<t})$). If retrieval occurs ($\mathbf{q}_t \neq \emptyset$), the generation step uses the extracted notes:
    \begin{equation}
    \begin{split}
     \mathbf{s}_t = & \text{LM}([\mathbf{O}_t, \mathbf{x}, \mathbf{s}_{<t}]) \\
                    & \text{where } \mathbf{O}_t \text{ is derived from } \mathcal{D}_{\mathbf{q}_t}
    \end{split}
    \end{equation}
    If retrieval is skipped ($\mathbf{q}_t = \emptyset$), $\mathbf{O}_t = \emptyset$ and generation proceeds without new retrieved context.
% Examples for each integration method are provided in \Cref{appx:examples}. 

%% file: latex/4Results.tex
\begin{table}[t] %[ht!]
\centering
% \small
\resizebox{0.9\linewidth}{!}{%
\begin{tabular}{@{}llc@{}}
\toprule
\textbf{Model} & \textbf{Dataset} & \textbf{Over Limit (n / \%)} \\
\midrule
\multirow{2}{*}{GPT-4o-mini (128k)} 
  & Frames      & 463 / 549  (84.3\%) \\
  & FanoutQA  & 244 / 310 (78.7\%) \\
\midrule
\multirow{2}{*}{LLaMA 3.1 70B (64K)} 
  & Frames      & 488 / 549 (88.8 \%) \\
  & FanoutQA   & 255 / 310 (82.2\%) \\
\bottomrule
\end{tabular}
}
\caption{Number of questions exceeding LLMs context with the top-5 Wikipedia pages (markdown format) being inserted into the LLM context at each step.}
\label{tab:context_limit}
\end{table}

\section{Experiments}
\input{latex/4_fanout-frames-table}

\label{sec:experiments}
\subsection{Datasets}
(1) \textbf{FanoutQA}~\cite{zhu-etal-2024-fanoutqa} focuses on "fanout" multi-hop, multi-document complex questions that require gathering information about a large set of entities. We report results on the dev set containing $310$ questions.\\
(2) \textbf{FRAMES}~\cite{krishna2024fact} a challenging multi-hop QA dataset requiring 2–15 hops to answer questions. We exclude questions requiring tabular reasoning and evaluate on $549$ examples. \\
(3) \textbf{HotpotQA}~\cite{yang-etal-2018-hotpotqa} a popular multi-hop QA dataset that requires reasoning over 2-3 Wikipedia articles. reasoning. We report results on $500$ examples from the dev set.\\ 
(4) \textbf{MultiHop-RAG}~\cite{tang2024multihop} a non-wikipedia based benchmark that involves retrieval over recent news articles. It has $\sim600$ news articles. For each question, we used BM25 to get the top five news articles in each iteration.

\paragraph{Evaluation metrics.} We report the \textbf{F1 score} between predicted and ground truth answer and following~\cite{krishna2024fact} we also use \textbf{GPT4-as-Judge score} with prompt in the appendix~\ref{fig:judge_prompt}. We also measure the \textbf{\emph{effective context length}} by reporting the average number of input \& output tokens processed by the main LLM and notes writing $\text{LM}_{\text{notes}}$ across all steps/iterations. Finally, we look at the \textbf{average number of steps} that is defined as the number of search queries that the main LLM needs to answer the question. 

\noindent
% \todo{discuss also experiment with secondary model being a smaller model}
\subsection{Models}
\label{sec:models}
We experiment with two LLMs, representing closed \& open weights, \texttt{GPT-4o-mini}\footnote{\url{https://openai.com/index/gpt-4o-mini-advancing-cost-efficient-intelligence/}} and \texttt{Llama 3.1-70-Instruct}~\cite{dubey2024llama}. We set the temperature to $0.7$ and use the same LLM for generating reasoning step and \noteswriting~(i.e LM = $\text{LM}_{\text{notes}}$). \texttt{Llama 3.1-70-Instruct} was  hosted using vLLM~\cite{kwon2023efficient} across $8$ A100-80GB GPUs, supporting a maximum context length of $64K$. \texttt{GPT-4o-mini}, which has a context length of $128K$.

\subsection{\noteswriting\ Implementation Details} 
% (will be moved here from 3.4)
For~\noteswriting, we utilize the Wikipedia API to fetch the top 5 relevant pages based on the query $\mathbf{q}_t$. Each retrieved Wikipedia page is converted to Markdown format using  \texttt{markdownify}\footnote{\url{https://pypi.org/project/markdownify/}} before being processed by $\text{LM}_{\text{notes}}$.

\noindent Initial experiments revealed that feeding the full content of the top-5 retrieved Wikipedia pages directly into the main LM (as standard iterative RAG baselines do) frequently caused context length errors, especially on challenging benchmarks like Fanout-QA and FRAMES. \Cref{tab:context_limit} shows that approximately 80\% of the questions are unanswerable as the context builds up and exceeds the context window. 
% \textit{Specifically, using GPT-4o-mini (128k context) on FRAMES, 463 out of 549 questions exceeded the context limit, and on Fanout-QA, 244 out of 310 questions failed similarly.} 
This observation, corroborated by the large average token counts reported for baselines Tables~\ref{tab:react_summary}, \ref{tab:ircot_summary}, and \ref{tab:flare_summary} (which often exceed typical LLM context limits, necessitating adjustments. Therefore, for a fair comparison, the baseline methods (IRCoT, FLARE, \react\ without \noteswriting) were re-implemented using a chunked document setup as detailed in Section~\ref{sec:baseline_details}, while \noteswriting\ operates on the full retrieved pages.

\subsection{Baseline Implementation Details}
\label{sec:baseline_details}
% (discusses differences for baselines)
\textbf{\react} --- As simply passing the top retrieved documents at each step to LLM causes context length exceeded (refer \Cref{tab:context_limit}), we re-implement the original baseline~\cite{yao2023react} which allows the LLM to \texttt{search} that retrieves the first paragraph of the top 5 Wikipedia pages, \texttt{select} that allows \react\ to select relevant page for first 10 passages and  \texttt{lookup} that returns paragraphs containing that specific string in the selected page.\\
\textbf{IRCoT \& FLARE} --- These were originally designed for older completion-based models such as \texttt{text-davinci-003} which relied heavily on internal parametric knowledge to generate CoTs. However, such a design is not directly applicable to questions requiring step by step planning and up-to-date knowledge. To address this, we adapt the few-shot prompting strategy to be compatible with chat models, enabling them not only to generate CoTs but also to explicitly plan ahead (see \Cref{appx:examples}). Moreover, the original baselines used BM25 from an older Wikipedia dump. However, in  initial experiments we observed that the older dump is outdated for latest datasets. Therefore, we used a recent dump \texttt{20231101.en}\footnote{\url{https://huggingface.co/datasets/wikimedia/wikipedia}} and dense passage retrieval with \texttt{ef-base-v2} embeddings~\cite{wang2022text}. We set the selective retrieval parameter $\theta$ to $0.8$ for all our experiments.\footnote{We also compare \noteswriting\ with Infogent~\cite{reddy2024infogent} with details and results in \Cref{appx:infogent}}

% \subsection{Baselines}
% \label{sec:baselines}

% \paragraph{ReAcT} For ReAcT, we used the the same fewshot prompt from the paper and used Wikipedia API to retrieve the wikipedia page with \texttt{lookup} as a tool that enables LMs find specifc keywords from the page. The fewshot prompt is in Appendix~\ref{fig:thought_action_prompt}.

\section{Results}  
\label{sec:results}
\paragraph{Enhanced performance with \noteswriting.} From Table~\ref{tab:react_summary},~\ref{tab:ircot_results} and~\ref{tab:flare_results} in comparison to the respective baselines, \noteswriting\  shows significant improvements across all models and benchmarks. Specifically from Table~\ref{tab:react_summary}, on complex long-form multihop-QA datasets like \textit{FRAMES} and \textit{Fanout-QA}, on average \renact\ achieves an absolute improvement of 29.1 points in F1 score and 21.1 points in GPT-4 score. On relatively easier datasets such as \textit{Hotpot-QA} and \textit{MultiHop-RAG}, ReNAcT yields absolute improvements of 10.3 and 5.0 points, respectively. The strong results compared to the baseline demonstrate that the LLM is receiving correct and relevant information at each step with \noteswriting.

From Tables~\ref{tab:ircot_results} and~\ref{tab:flare_results} on challenging datasets \noteswriting\ coupled with each of IRCoT and FLARE leads to 14.4 and 10.5 points improvement on F1 and GPT-4 score. Similarly on Hotpot-QA and MultiHop-RAG we find 7.0 and 10.8 points improvement on F1 and GPT-4 score respectively. 
% Additional results are available in Appendix~~\ref{tab:ircot_summary} \&~\ref{tab:flare_summary}.
\begin{table}[t]
\centering
% \small
\resizebox{0.88\linewidth}{!}{%
\begin{tabular}{@{}lllcc@{}}
\toprule
\textbf{Model} & \textbf{Setting} & \textbf{Benchmark} & \textbf{F1 (\%)} & \textbf{GPT-4 (\%)} \\
\midrule

\multirow{8}{*}{\rotatebox[origin=c]{90}{\textbf{GPT-4o-mini}}}
& \multirow{4}{*}{IRCoT}
& FanoutQA     & 33.6 & 15.2 \\
&               & Frames        & 24.0 & 22.0 \\
&               & HotpotQA     & 36.4 & 42.8 \\
&               & M-RAG  & 23.7 & 48.0 \\
\cmidrule{2-5}
& \multirow{4}{*}
& FanoutQA     & \textbf{41.9} & \textbf{21.3} \\
 &       IRCoT        & Frames        & \textbf{43.9} & \textbf{42.3} \\
 &       +\textit{Notes}        & HotpotQA     & \textbf{46.2} & \textbf{53.8} \\
&        \textit{Writing}        & M-RAG  & \textbf{36.0} & \textbf{65.6} \\

\midrule

\multirow{8}{*}{\rotatebox[origin=c]{90}{\textbf{LLaMA-3.1-70B}}}
& \multirow{4}{*}{IRCoT}
& FanoutQA     & 21.0 & 8.4 \\
&               & Frames        & 19.4 & 21.1 \\
&               & HotpotQA     & 31.5 & 38.2 \\
&               & M-RAG  & 35.4 & 64.8 \\
\cmidrule{2-5}
& \multirow{4}{*}{}
& FanoutQA     & \textbf{36.0} & \textbf{22.9} \\
&   IRCoT            & Frames        & \textbf{26.9} & \textbf{33.3} \\
&   +\textit{Notes}            & HotpotQA     & \textbf{36.5} & \textbf{53.0} \\
&      \textit{Writing}         & M-RAG  & \textbf{38.0} & \textbf{64.8} \\

\bottomrule
\end{tabular}
}
\caption{IRCoT performance for GPT-4o-mini and LLaMA-3.1-70B M-RAG represents MultiHop-RAG.}
\label{tab:ircot_results}
\end{table}

\paragraph{Increased effective context length.} 
% \Cref{tab:react_summary,tab:ircot_results,tab:flare_results}
Tables~\ref{tab:react_summary},~\ref{tab:ircot_summary} and~\ref{tab:flare_summary} 
show the average number of input and output tokens across all steps for the baseline and \noteswriting. The total number of tokens processed by the system (sum of input tokens across main and notes-writing LLMs) increases, allowing the model to reason over more retrieved content. However, it is important to note that this information cannot be naively appended to the main LLM's context (summing columns 7 and 9 would exceed the context window). This demonstrates that \noteswriting\ enables scalable use of large retrieval context by delegating information management to a specialized LLM.

With ReAct (\Cref{tab:react_summary}),
% For Table~\ref{tab:react_summary}  
the number of tokens for the main LLM reduce significantly from baseline to \renact\ across all benchmarks — by 77K tokens for GPT-4o-mini and 86K tokens for LLaMA-3.1-70B on average. This demonstrates concise notes being added at each retrieval step. Similarly, output tokens decrease across benchmarks, with an average reduction of 96 tokens for GPT-4o-mini and 53 tokens for LLaMA-3.1-70B. The same trend is observed with IRCoT (\Cref{tab:ircot_summary}) and FLARE (\Cref{tab:flare_summary}) where the main LLM input tokens reduces by at least 4x and 1.5x for GPT-4o-mini and LLaMA-3.1-70B, with output tokens being almost comparable.

\paragraph{Reduced average steps.} 
% Table~\ref{tab:react_summary} shows that with \renact, 
\renact~reduces the average number of steps across all the benchmarks \textit{MultiHop-RAG} (\Cref{tab:react_summary}). For \textit{Frames} and \textit{Fanout-QA}, the reduction is 2.23 \& 1.86 for GPT-4o-mini and LLaMA-3.1-70B respectively. The reduction is smaller but still present for \textit{Hotpot-QA} with an average drop of 0.13 and 0.96 steps respectively. We further analyze the reduction in redundant queries and correlation with ground truth steps in Section~\ref{sec:analysis}.

\paragraph{\noteswriting~is cost effective.} Tables~\ref{tab:react_summary},~\ref{tab:ircot_summary}, and~\ref{tab:flare_summary} show that with~\noteswriting\ the combined output tokens by the main \& note taking LLM are on an average 2-3x more than the baselines. However, this tradeoff is justified by the performance improvement and the output tokens being significantly less (about 100x) than effective number of input tokens. As the output tokens are the major contributing factor to cost~\footnote{https://openai.com/api/pricing/} \& latency, \noteswriting\ is a much more cost + compute + performance effective approach.

%% file: latex/4_fanout-frames-table.tex
\begin{table*}[h!]
\centering
% \small
\resizebox{0.8\linewidth}{!}{%
\begin{tabular}{lclccccccc}
\toprule
\textbf{Model} & \textbf{Setting} & \textbf{Benchmark} & \textbf{F1 (\%)} & \textbf{GPT-4} & \textbf{Avg}
& \multicolumn{2}{c}{\textbf{Main Tokens}} & \multicolumn{2}{c}{\textbf{Notes Tokens}} \\
\cmidrule(lr){7-8} \cmidrule(lr){9-10}
& & & & \textbf{Score} (\%) & \textbf{steps} & Input & Output & Input & Output \\
\midrule

\multirow{8}{*}{\rotatebox[origin=c]{90}{\textbf{GPT-4o-mini}}}
& \multirow{4}{*}{\react}
& Fanout-QA      & 28.6 & 12.9 & 10.4 & 116K & 916  & -     & -    \\
&                & Frames        & 8.7 & 31.1 & 8.36 & 67K  & 707  & -     & -    \\
&                & Hotpot-QA     & 42.2 & 56.4 & 3.33 & 26K  & \textbf{319}  & -     & -    \\
&                & MultiHop-RAG  & \textbf{58.0} & 64.2 & \textbf{5.6} & 188K  & \textbf{278}  & -     & -    \\
\cmidrule{2-10}
& \multirow{4}{*}
& Fanout-QA      & \textbf{50.0} & \textbf{28.0} & \textbf{7.8} & \textbf{17K}  & \textbf{598}  & 359K  & 675  \\
&      \react\ +          & Frames        & \textbf{46.8} & \textbf{52.3} & \textbf{6.51} & \textbf{16K}  & \textbf{543}  & 277K  & 607  \\
&       \noteswriting\         & Hotpot-QA     & \textbf{51.0} & \textbf{64.0} & \textbf{3.2} & \textbf{9K}   & 326  & 130K  & 321  \\
& (\renact)          & MultiHop-RAG  & \textbf{58.0} & \textbf{70.6} & 6.0 & \textbf{46K}  & 368  & 68K   & 390  \\

\midrule

\multirow{8}{*}{\rotatebox[origin=c]{90}{\textbf{LLaMA-3.1-70B}}}
& \multirow{4}{*}{\react}
& Fanout-QA      & 13.5 & 8.7 & 7.5 & 113K   & 506    & -     & -    \\
&                & Frames        & 21.7 & 26.8 & 6.83 & 85K    & 433    & -     & -    \\
&                & Hotpot-QA     & 43.7 & 52.6 & 4.3 & 49K    & 289    & -     & -    \\
&                & MultiHop-RAG  & 53.6 & 61.4 & \textbf{5.24} & 180K  & 295  & -     & -    \\
\cmidrule{2-10}
& \multirow{4}{*}
& Fanout-QA      & \textbf{43.0}    & \textbf{26.1} & \textbf{5.80}    & \textbf{15K}    & \textbf{485}    & 265K     & 1116    \\
&        \react +        & Frames        & \textbf{49.0}    & \textbf{57.6} & \textbf{4.81}    & \textbf{13K}    & \textbf{412}   & 193K     & 717   \\
&       \noteswriting\         & Hotpot-QA     & \textbf{55.5}    & \textbf{67.4} & \textbf{3.34}    & \textbf{8K}    & \textbf{274}    & 109K     & 391    \\
&   (\renact)             & MultiHop-RAG  & \textbf{63.5} & \textbf{73.0} & 5.9 & \textbf{47K}  & \textbf{262}  & {76K}   & {425}  \\

\bottomrule
\end{tabular}
}
\caption{\react\ and \noteswriting\ results for GPT-4o-mini and LLaMA. Main tokens represent the total number of input \& output tokens for the main LLM across all steps (average on all questions). Similarly, notes tokens represent the total number of input \& output tokens across all steps by the notes writing LLM (averaged on all questions). Token counts are rounded to the nearest thousand.}
\vspace{-1mm}
\label{tab:react_summary}
\end{table*}

%% file: latex/5_additional_results.tex
% \label{sec:additional_resuls}

% \subsection{IRCoT \& FLARE results} 
% \label{sec:ircot_flare}

% Tables~\ref{tab:ircot_results} and~\ref{tab:flare_results} shows the performance improvements with \noteswriting\ on the IRCoT and FLARE baselines. On challenging datasets like FRAMES and Fanout-QA \noteswriting\ leads to 14.4 and 10.5\% improvement on F1 and GPT-4 score. Similarly on Hotpot-QA and MultiHop-RAG we find 7.0 and 10.8\% improvement on F1 and GPT-4 score respectively. Additional results are available in Appendix~~\ref{tab:ircot_summary} \&~\ref{tab:flare_summary}.

% \paragraph{FLARE} Like IRCoT, FLARE was originally tailored for completion-based models and thus also depends on internal knowledge to generate CoTs. Consequently, we revise FLARE’s prompting strategy to ensure compatibility with chat models, preserving both selective retrieval and reasoning capabilities. Similar to IRCoT we used the above wikipedia dump for FLARE and used dense passage retrieval with \texttt{ef-base-v2}~\cite{wang2022text}. We set $\theta$ to $0.8$ for all our experiments. 

\begin{table}[t]
\centering
% \small
\resizebox{0.88\linewidth}{!}{%
\begin{tabular}{@{}lllcc@{}}
\toprule
\textbf{Model} & \textbf{Setting} & \textbf{Benchmark} & \textbf{F1 (\%)} & \textbf{GPT-4 (\%)} \\
\midrule

\multirow{8}{*}{\rotatebox[origin=c]{90}{\textbf{GPT-4o-mini}}}
& \multirow{4}{*}{FLARE}
& Fanout-QA     & 35.1 & 14.2 \\
&               & Frames        & 26.3 & 23.7 \\
&               & HotpotQA     & 34.8 & 39.0 \\
&               & M-RAG  & 28.9 & 65.7 \\
\cmidrule{2-5}
& \multirow{4}{*}{}
& Fanout-QA     & \textbf{42.3} & \textbf{22.2} \\
&     FLARE          & Frames        & \textbf{27.7} & \textbf{29.8} \\
&     +\textit{Notes}          & HotpotQA     & \textbf{34.5} & \textbf{45.8} \\
&     \textit{Writing}           & M-RAG  & \textbf{30.2} &\textbf{66.6} \\

\midrule

\multirow{8}{*}{\rotatebox[origin=c]{90}{\textbf{LLaMA-3.1-70B}}}
& \multirow{4}{*}{FLARE}
& FanoutQA     & 23.0 & 11.4 \\
&               & Frames        & 16.4 & 18.6 \\
&               & Hotpot-QA     & 24.7 & 31.2 \\
&               & M-RAG  & 36.1 & 67.0 \\
\cmidrule{2-5}
& \multirow{4}{*}{}
& FanoutQA     & \textbf{35.8} & \textbf{24.2} \\
&     FLARE          & Frames        & \textbf{20.0} & \textbf{25.3} \\
&     +\textit{Notes}          & HotpotQA     & \textbf{34.0} & \textbf{47.0} \\
&      \textit{Writing}         & M-RAG  & \textbf{30.5} & \textbf{66.4} \\

\bottomrule
\end{tabular}
}
\caption{FLARE performance for GPT-4o-mini and LLaMA-3.1-70B. M-RAG represents MultiHop-RAG.}
\label{tab:flare_results}
\end{table}

\begin{figure*}[t]
\centering
\small
 \includegraphics[height=0.40\linewidth,width=0.9\linewidth]{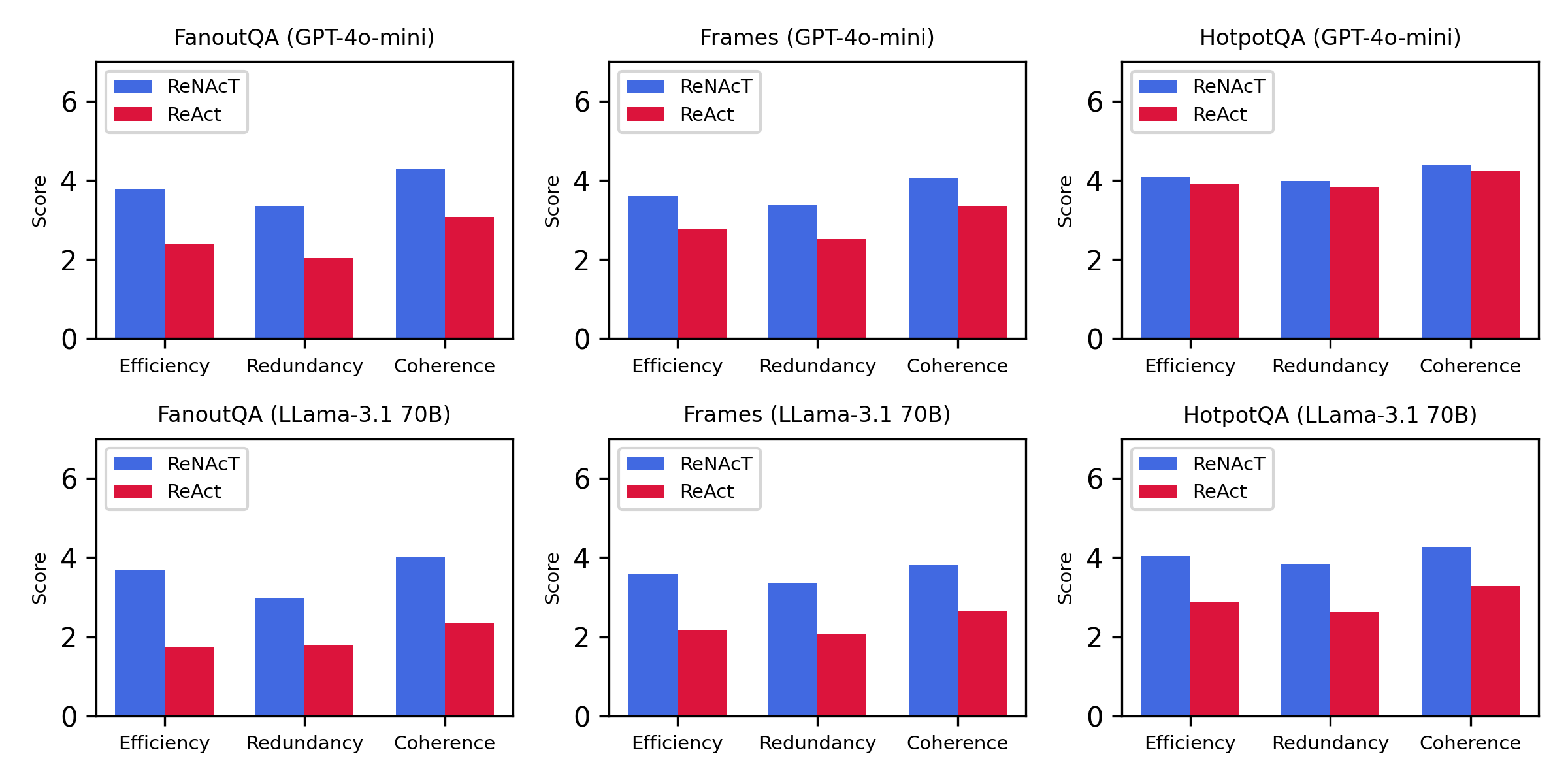} % Ensure path is correct
 \caption{Quality evaluation of \react\ and \renact\ reasoning chain.}
 \label{fig:quality_eval}
\end{figure*}

\section{Analysis} \label{sec:analysis}

\paragraph{Reasoning Quality Analysis.}
We evaluate the reasoning chains generated by \react\ and \renact\ using GPT-4o as a judge across three axes, (1) Efficiency --- to measure redundant searches and how well each step contributes to the final answer, (2) Redundancy --- to assess repeated search queries, or unnecessary repetition or duplication of steps (3) Coherence --- to check if the chain is comprehensible, logically connected, and free from unnecessary complexity or ambiguity. The evaluation prompt is in Appendix~\ref{fig:quality_prompt}. Figure~\ref{fig:quality_eval} shows the results. \renact\ is better across all three axes than \react\ across on all models and datasets. Specifically, on Frames and FanoutQA across both models efficiency, redundancy and coherence improve by at least 1.5x. On HotpotQA, the improvement is 1.2x. 

\begin{figure*}[t]
\centering
\small
 \includegraphics[height=0.45\linewidth,width=0.9\linewidth]{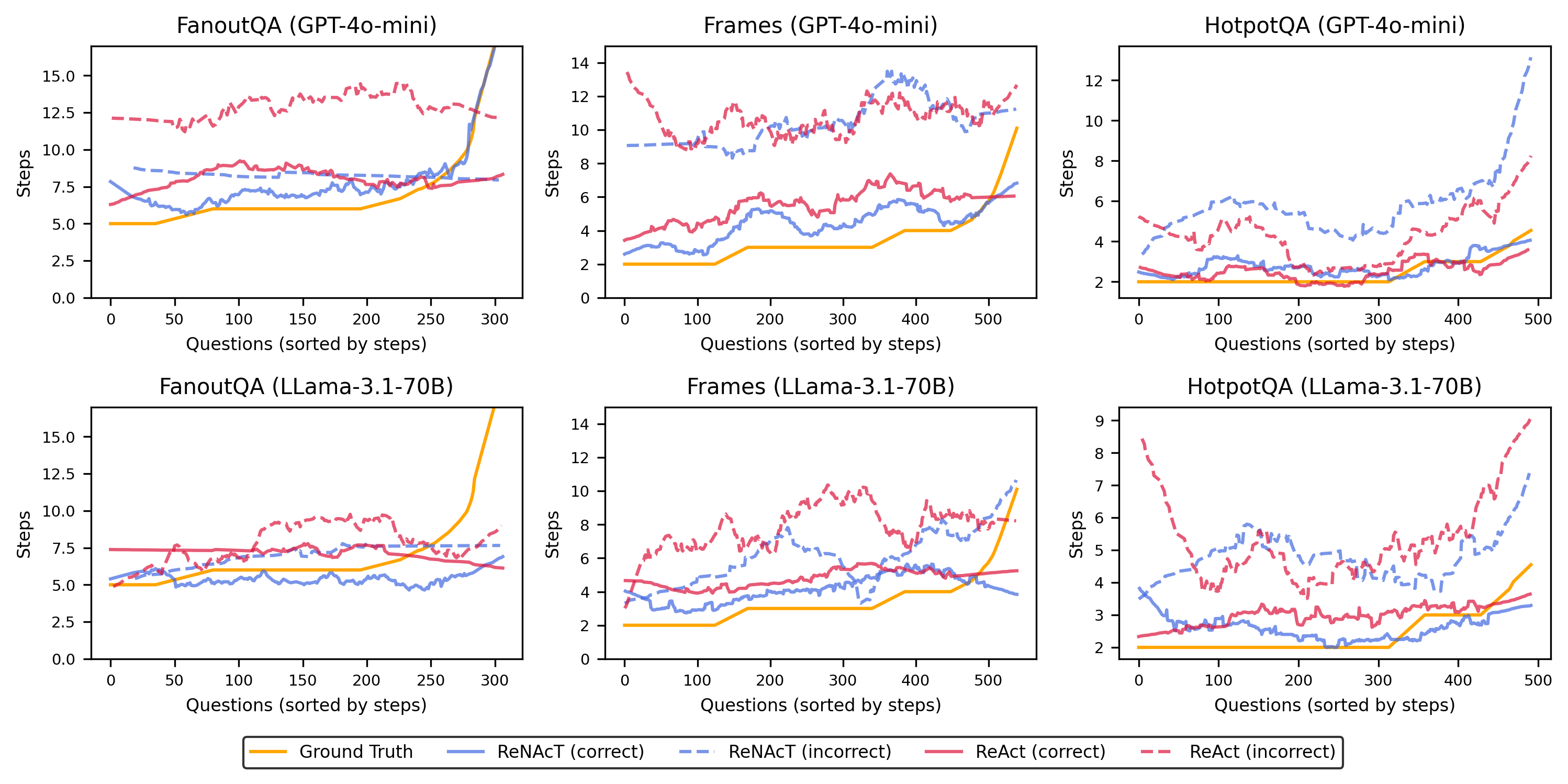} % Ensure path is correct
 \caption{Steps (smoothed) by \renact, \react\ vs the ground truth steps for GPT-4o-mini and LLama-3.1-70B.}
 \label{fig:steps}
\end{figure*}

\begin{table}[t]
\centering
% \small
\resizebox{0.7\linewidth}{!}{%
\begin{tabular}{@{}llcc@{}}
\toprule
% \multirow{2}{*}{Model} & \multirow{2}{*}{Dataset} & \multicolumn{2}{c}{\%} \\
%                        &                          & NotesWriting & ReAct \\
Model & Dataset & \renact\ & \react \\
\midrule
\multirow{3}{*}{GPT-4o-mini} 
    & Fanout           &\textbf{22.83} & 36.84 \\
    & Frames           & \textbf{23.18} & 29.89 \\
    & HotpotQA         & \textbf{30.56} & 46.02 \\
\midrule
\multirow{3}{*}{LLaMA}  
    & Fanout           & 49.44 & \textbf{35.71} \\
    & Frames           & \textbf{28.22} & 30.94 \\
    & HotpotQA         & \textbf{31.33} & 32.82 \\
\bottomrule
\end{tabular}
}
\caption{\% of correct questions having search steps less than number of ground truth Wikipedia pages.}
\label{tab:false_positives}
\end{table}
\vspace{-2mm}
\paragraph{Search Steps Comparison.}

\Cref{fig:steps} shows the comparison of the number of ground truth steps, \react\ and \renact\ search steps for each question in each dataset across both models. The dashed lines for each method represents the in-correct answers and the sold line represents correct ones. The x-axis is the index of the question in the dataset sorted by the number of ground-truth search steps. From the \Cref{fig:steps}, it can be observed that \renact\ (solid blue line) is much closer to the ground truth steps with \react\ (solid red line) being relatively far demonstrating the effectiveness of \noteswriting\ in coming up with correct stepwise plan and search query for retrieval. \Cref{fig:steps} also shows that the in-correct questions (dashed red \& blue line) have a higher number of steps that shows that it fails after many re-tries. The difference among \renact\, \react\ and ground truth steps is more significant in challenging datasets like Frames and Fanout-QA as opposed to HotpotQA.

\paragraph{Correct Answers with Fewer Searches than Ground Truth.}
\Cref{fig:steps} shows cases where \renact\ and \react\ (solid blue \& red lines) are below ground truth steps. \Cref{tab:false_positives} reports percentages of cases for
% We report the percentages of such cases in \Cref{tab:false_positives}. This shows 
the number of correctly answered questions which took less searches than the number of ground truth Wikipedia pages required to answer the question correctly.

% \subsection{Performance on Easy vs Hard examples}
% \begin{table}[htbp!]
%     \centering
%     \resizebox{\columnwidth}{!}{
%     \begin{tabular}{cccc}
%     \toprule
%     Model & Dataset & \reactbm & \renact \\ \midrule
%     \multirow{2}{*}{LLama-3.1-70B-Inst} & FanoutQA & 7.43 & \textbf{2.01} \\ \cmidrule(lr){2-4}
%      & FRAMES & 5.12 & \textbf{2.20} \\ \midrule
%     \multirow{2}{*}{GPT-4o-mini} & FanoutQA & 8.03 & \textbf{2.69} \\ \cmidrule(lr){2-4}
%      & FRAMES & 5.06 & \textbf{3.23} \\ \bottomrule
%     \end{tabular}%
%     }
%     \caption{\footnotesize No of search retries are lower for \renact compared to \reactbm~for FanoutQA and FRAMES.}
%     \label{tab:retries}
%     \vspace{-4mm}
% \end{table}

%% file: latex/6_Conclusion.tex
\section{Conclusion} \label{sec:conclusion}

We presented \noteswriting, a plug-and-play module that improves \emph{effective context length} in iterative RAG by accumulating only the most relevant information at each reasoning step. Experiments on \emph{three} RAG baselines (IRCoT, FLARE, and \react), \emph{four} multi-hop QA datasets, and \emph{two} LLMs show that \noteswriting\ improves performance by up to 15.6 points, while also reducing context overload, the number of reasoning steps, and redundancy. In the \react\ setting, \noteswriting\ enables better planning by guiding the model to generate more accurate search queries and retrieve the correct documents. Moreover, \noteswriting\ consistently improves coherence and efficiency of planning and search across models in \react. Therefore, we suggest \renact\ as an effective iterative RAG framework. Our results show that \renact\ (\react\ + \noteswriting) makes iterative RAG more scalable and precise.

\section*{Limitations and Societal Impact} \label{sec:limitations}

Our approach has several limitations. First, our experiments are limited to the two models we experiment with, which could be extended to newer smaller open-source models. Second, we limit online searches to the Wikipedia API\footnote{\url{https://www.mediawiki.org/wiki/API:Search}}, which only supports searching for text matching Wiki pages; and third, Wiki pages change often and this could lead to a mismatch with static benchmarks' ground truth. While these could affect performance, we ensure that the same setup is also followed in all baselines we experiment with, to keep evaluation comparable while reducing the need to utilize paid search APIs. Third, with retrievals based on iterative notes writing, there is a possibility of conflicting information being received (\Cref{tab:conflict_example_frames}). It is possible that the model starts hallucinating facts, and this remains a weakness at large. Lastly, we impose a maximum iteration limit to ensure computational efficiency, which could also impact performance. Further explorations towards improving on weaknesses remain future work.

Potential risks of our work include usage in scenarios where the requested retrieval information is toxic or harmful. While we cannot control how our method is used for prompting, we expect content moderation policies to help with reducing the impact of such queries. Moreover, hallucinations~\ref{tab:hallucinated_example_1},~\ref{tab:hallucinated_example_2} can affect the QA experience, although manual observation of the reasoning traces show that recovery can be better with \noteswriting.

We expect our work to significantly enhance the QA user experience, as focused information improves performance and reduced context lengths lower computational costs. We hope our \noteswriting~method can contribute towards better task handling at large. We will make our code publicly available upon acceptance towards this goal.

%% file: latex/7_Appendix.tex
\pagebreak
\section{Appendix}

\subsection{Infogent Implementation Details}
\label{appx:infogent}
We use the official code provided by Infogent authors \href{https://github.com/gangiswag/infogent/tree/main/direct-api-driven}{here} (Apache 2.0. License) with the following modifications:
\begin{itemize} [noitemsep, nolistsep]
    \item Due to our limitations in accessing OpenAI, we modify the code to use AzureOpenAI.
    \item OpenAI embedding is replaced by sentence-transformers' \texttt{all-mpnet-base-v2}\footnote{\url{https://huggingface.co/sentence-transformers/all-mpnet-base-v2}}.
    \item Serper Google Search\footnote{\url{https://serper.dev/}} is replaced by Wikipedia search API due to credit limitations and to use similar open knowledge tools as those used in our method, reducing the cost needed to conduct RAG experiments. 
    % \shiva{Should we also mention how black box google search might obfuscate true search cost?}
\end{itemize}
\subsection{Results}
\begin{table}[ht]
\centering
\small
\begin{tabular}{llcc}
\toprule
\textbf{Setting} & \textbf{Benchmark} & \textbf{F1 (\%)} & \textbf{GPT-4 (\%)} \\
\midrule

 \multirow{2}{*}{InfoAgent}
& FanoutQA     & 47.2 & 22.9 \\
         & Frames        & 28.0 & 29.9 \\
\midrule
\multirow{2}{*}{\noteswriting}
& FanoutQA     & \textbf{50.0} & \textbf{28.0} \\
         & Frames        & \textbf{46.8} & \textbf{52.3} \\

\bottomrule
\end{tabular}
\caption{Infoagent vs \noteswriting~performance comparison on GPT-4o-mini.}
\label{tab:infoagent_results}
\end{table}

\subsection{Benchmarks}
\label{appx:benchmarks}
We evaluated four multi-hop QA datasets: (1) FanOutQA~\cite{zhu-etal-2024-fanoutqa}, which features complex fanout questions, (2) FRAMES~\cite{krishna2024fact}, requiring reasoning over 2–15 articles, (3) MultiHop-RAG~\cite{tang2024multihop}, which involves retrieval and reasoning over news articles, and (4) HotpotQA~\cite{yang-etal-2018-hotpotqa}, which requires multi-article reasoning. For FanOutQA, we evaluated all $310$ examples from the development set, while for FRAMES, we used $549$ multiple-constraint-tagged questions. For MultiHop-RAG and HotpotQA, we assessed performance on $500$ examples from the test and development splits, respectively. FanOutQA, HotpotQA and Wikipedia comes under CC BY-SA 4.0 (Creative Commons Attribution-ShareAlike 4.0 International License), FRAMES under Apache 2.0. license and MultiHop-RAG under ODC-By (Open Data Commons Attribution License).

\subsection{Models}
\label{appx:models}
\textbf{Models}: We conduct experiments with two models, representing both closed and open weights: \texttt{GPT-4o-mini}\footnote{\url{https://openai.com/index/gpt-4o-mini-advancing-cost-efficient-intelligence/}} and \texttt{Llama 3.1-70-Instruct}~\cite{dubey2024llama}. The temperature is set to $0.7$, and the same LLM is used for generating reasoning steps and \noteswriting~(i.e., $\mathcal{M}_s = \mathcal{M}$). \texttt{Llama 3.1-70-Instruct} was  hosted using vLLM~\cite{kwon2023efficient} across $8$ A100-80GB GPUs, supporting a maximum context length of $64K$. With parallelization, evaluation runs took approximately $9$–$10$ hours for MultiHop-RAG, HotpotQA, and FRAMES, and around $15$ hours for FanOutQA. \texttt{GPT-4o-mini}, which has a context length of $128K$, completed evaluations in approximately $7$ hours for FRAMES and FanOutQA, $2$ hours for HotpotQA, and $27$ minutes for MultiHop-RAG. The reported times include the full end-to-end process, accounting for rate limits, Wikipedia queries, and \noteswriting.

\subsection{Standard deviation across runs}
\label{appx:stddev}
We ran the \noteswriting~ and \renact~ across all datasets and models three times to see the variance across different runs. We report the results in Table~\ref{tab:std_metrics}.
\begin{table}[h!]
    \centering
    \small
    \begin{tabular}{cccc}
        \toprule
        \textbf{Model} & \textbf{Dataset} & \textbf{Avg F1} & \textbf{GPT-4 Score} \\
        \cmidrule{1-4}
        \multirow{2}{*}{GPT-4o-mini} & Fanout  & $\pm$ 1.86  & $\pm$ 2.45  \\
        & Frames  & $\pm$ 1.10  & $\pm$ 2.35 \\
        \cmidrule{1-4}
        \multirow{2}{*}{Llama-3.1 70B} & Fanout  & $\pm$ 3.79  & $\pm$ 1.54 \\
        & Frames  & $\pm$ 4.42  & $\pm$ 5.76  \\
        \bottomrule
    \end{tabular}
    \caption{Standard deviation across Frames \& FanoutQA.}
    \label{tab:std_metrics}
\end{table}
\noindent 

\begin{table*}[ht]
\centering
\small
\begin{tabular}{lcccccccc}
\toprule
\textbf{Model} & \textbf{Setting} & \textbf{Benchmark} & \textbf{F1 (\%)} & \textbf{GPT-4} 
& \multicolumn{2}{c}{\textbf{Main Tokens}} & \multicolumn{2}{c}{\textbf{Notes Tokens}} \\
\cmidrule(lr){6-7} \cmidrule(lr){8-9}
& & & & Score (\%) & Input & Output & Input & Output \\
\midrule

\multirow{8}{*}{\rotatebox[origin=c]{90}{\textbf{GPT-4o-mini}}}
& \multirow{4}{*}{Baseline}
& Fanout-QA      & 33.6 & 15.2 & 273K & 385  & -     & -    \\
&                & Frames        & 24.0 & 22.0 & 183K & 312  & -     & -    \\
&                & Hotpot-QA     & 36.4 & 42.8 & 99K  & 205  & -     & -    \\
&                & MultiHop-RAG  & 23.7 & 48.0 & 909K & 322  & -     & -    \\
\cmidrule{2-9}
& \multirow{4}{*}{\noteswriting}
& Fanout-QA      & 41.9 & 21.3 & 68K  & 444  & 902K  & 1.8K \\
&                & Frames        & 43.9 & 42.3 & 37K  & 280  & 658K  & 862  \\
&                & Hotpot-QA     & 46.2 & 53.8 & 26K  & 193  & 433K  & 650  \\
&                & MultiHop-RAG  & 36.0 & 65.6 & 189K & 412 & 40K   & 324  \\

\midrule

\multirow{8}{*}{\rotatebox[origin=c]{90}{\textbf{LLaMA-3.1-70B}}}
& \multirow{4}{*}{Baseline}
& Fanout-QA      & 21.0 & 8.4  & 197K & 511  & -     & -    \\
&                & Frames        & 19.4 & 21.1 & 192K & 732  & -     & -    \\
&                & Hotpot-QA     & 31.5 & 38.2 & 167K & 557  & -     & -    \\
&                & MultiHop-RAG  & 35.4 & 64.8 & 429K & 313 & -     & -    \\
\cmidrule{2-9}
& \multirow{4}{*}{\noteswriting}
& Fanout-QA      & 36.0 & 22.9 & 154K & 552  & 1.4M  & 7.6K \\
&                & Frames        & 26.9 & 33.3 & 116K & 735  & 1.2M  & 3.9K \\
&                & Hotpot-QA     & 36.5 & 53.0 & 100K & 561  & 1.1M  & 3.3K \\
&                & MultiHop-RAG  & 38.0 & 64.8 & 99K  & 680  & 338K  & 2.9K \\

\bottomrule
\end{tabular}
\caption{IRCoT results for GPT-4o-mini and LLaMA-3.1-70B across Baseline and \noteswriting. Token counts are rounded to the nearest thousand (input/output).}
\label{tab:ircot_summary}
\end{table*}
\begin{table*}[ht]
\centering
\small
\begin{tabular}{lcccccccc}
\toprule
\textbf{Model} & \textbf{Setting} & \textbf{Benchmark} & \textbf{F1 (\%)} & \textbf{GPT-4} 
& \multicolumn{2}{c}{\textbf{Main Tokens}} & \multicolumn{2}{c}{\textbf{Notes Tokens}} \\
\cmidrule(lr){6-7} \cmidrule(lr){8-9}
& & & & Score (\%) & Input & Output & Input & Output \\
\midrule

\multirow{8}{*}{\rotatebox[origin=c]{90}{\textbf{GPT-4o-mini}}}
& \multirow{4}{*}{Baseline}
& Fanout-QA      & 35.1 & 14.2 & 268K & 384  & -     & -    \\
&                & Frames        & 26.3 & 23.7 & 50K  & 314  & -     & -    \\
&                & Hotpot-QA     & 34.8 & 39.0 & 36K  & 218  & -     & -    \\
&                & MultiHop-RAG  & 28.9 & 65.7 & 204K & 319  & -     & -    \\
\cmidrule{2-9}
& \multirow{4}{*}{\noteswriting}
& Fanout-QA      & 42.3 & 22.2 & 65K  & 409  & 968K  & 2.2K \\
&                & Frames        & 27.7 & 29.8 & 43K  & 358  & 47K   & 56   \\
&                & Hotpot-QA     & 34.5 & 45.8 & 33K  & 258  & 45K   & 49   \\
&                & MultiHop-RAG  & 30.2 & 66.6 & 205K  & 312  & 14K   & 49   \\

\midrule

\multirow{8}{*}{\rotatebox[origin=c]{90}{\textbf{LLaMA-3.1-70B}}}
& \multirow{4}{*}{Baseline}
& Fanout-QA      & 23.0    & 11.4    & 195K    & 453    & -     & -    \\
&                & Frames        & 16.4 & 18.6 & 86K  & 640  & -     & -    \\
&                & Hotpot-QA     & 24.7 & 31.2 & 83K  & 573  & -     & -    \\
&                & MultiHop-RAG  & 36.1    & 67.0    & 301K    & 558     & - & -    \\
\cmidrule{2-9}
& \multirow{4}{*}{\noteswriting}
& Fanout-QA      & 35.8 & 24.2 & 154K & 555  & 1.3M  & 7.5K \\
&                & Frames        & 20.0 & 25.3 & 67K  & 592  & 47K   & 66   \\
&                & Hotpot-QA     & 34.0 & 47.0 & 65K  & 563  & 45K   & 77   \\
&                & MultiHop-RAG  & 30.5 & 66.4 & 67K  & 710  & 15K   & 38   \\

\bottomrule
\end{tabular}
\caption{FLARE results for GPT-4o-mini and LLaMA-3.1-70B across Baseline and \noteswriting. Token counts are rounded to the nearest thousand (input/output).}
\label{tab:flare_summary}
\end{table*}

% \subsection{Comparing no of retries between~\renact~ and BM25}

\subsection{Examples comparing~\renact~with baselines}
\label{appx:examples}

\renewcommand{\arraystretch}{1.5}
\begin{table*}[htbp!]
    \centering
    \small
    \begin{tabular}{|p{\textwidth}|}
        \hline
        \textbf{Question:} What law schools did the most recent four Supreme Court justices attend?  \\
        \hline
        \textbf{Thought:} I need to gather information on the law schools attended by the most recent four Supreme Court justices. First, I will search for the names of the most recent four justices to find out where they studied law. \\
        \textbf{Action:} search[Supreme Court justices; Who are the most recent four Supreme Court justices?] \\
        \textbf{Observation:} (Result 1) List of justices of the Supreme Court of the United States - The most recent four Supreme Court justices are: Amy Coney Barrett, Ketanji Brown Jackson, Neil Gorsuch, and Brett Kavanaugh. \\
        (Result 2) Ideological leanings of United States Supreme Court justices - The most recent four Supreme Court justices are: Amy Coney Barrett (appointed in 2020), Ketanji Brown Jackson (appointed in 2022), Brett Kavanaugh (appointed in 2018), and Neil Gorsuch (appointed in 2017). \\
        (Result 3) Demographics of the Supreme Court of the United States - The most recent four Supreme Court justices are: Ketanji Brown Jackson (appointed in 2022), Amy Coney Barrett (appointed in 2020), Brett Kavanaugh (appointed in 2018), and Neil Gorsuch (appointed in 2017). \\
        \hline
        \textbf{Thought:} The most recent four Supreme Court justices are Amy Coney Barrett, Ketanji Brown Jackson, Neil Gorsuch, and Brett Kavanaugh. Next, I need to find out which law schools they attended. I will start by searching for the law school attended by Amy Coney Barrett. \\
        \textbf{Action:} search[Amy Coney Barrett; Which law school did Amy Coney Barrett attend?] \\
        \textbf{Observation:} (Result 1) Amy Coney Barrett attended Rhodes College (BA) and University of Notre Dame (JD). She graduated from Notre Dame Law School in 1997 with a Juris Doctor, \textit{summa cum laude}, ranked first in her class. \\
        \hline
        \textbf{Thought:} Amy Coney Barrett attended Notre Dame Law School. Next, I need to find out which law school Ketanji Brown Jackson attended. \\
        \textbf{Action:} search[Ketanji Brown Jackson; Which law school did Ketanji Brown Jackson attend?] \\
        \textbf{Observation:} (Result 1) Ketanji Brown Jackson attended Harvard Law School, where she graduated in 1996 with a Juris Doctor, \textit{cum laude}. \\
        (Result 2) Joe Biden Supreme Court candidates - Ketanji Brown Jackson attended Harvard University for her law degree. \\
        (Result 3) Demographics of the Supreme Court of the United States - Ketanji Brown Jackson attended Harvard Law School. \\
        \hline
        \textbf{Thought:} Ketanji Brown Jackson attended Harvard Law School. Now, I need to find out which law school Neil Gorsuch attended. \\
        \textbf{Action:} search[Neil Gorsuch; Which law school did Neil Gorsuch attend?] \\
        \textbf{Observation:} (Result 1) Neil Gorsuch received his legal education at Harvard Law School and graduated in 1991 with a Juris Doctor, \textit{cum laude}. \\
        \hline
        \textbf{Thought:} Neil Gorsuch also attended Harvard Law School. Finally, I need to find out which law school Brett Kavanaugh attended. \\
        \textbf{Action:} search[Brett Kavanaugh; Which law school did Brett Kavanaugh attend?] \\
        \textbf{Observation:} (Result 1) Brett Kavanaugh - Kavanaugh graduated from Yale Law School with a Juris Doctor degree in 1990. \\
        \hline
        \textbf{Thought:} Brett Kavanaugh graduated from Yale Law School. Now I have gathered the law schools attended by the most recent four Supreme Court justices: Amy Coney Barrett attended Notre Dame Law School, Ketanji Brown Jackson and Neil Gorsuch both attended Harvard Law School, and Brett Kavanaugh attended Yale Law School. \\
        \textbf{Action:} finish[Notre Dame Law School, Harvard Law School, Harvard Law School, Yale Law School] \\
        \hline
    \end{tabular}
    \caption{An example of~\noteswriting~on the FanoutQA dataset using GPT-4o-mini.}
    \label{tab:example_fanout}
\end{table*}

\renewcommand{\arraystretch}{1.5}
\begin{table*}[htbp!]
    \centering
    \small
    \begin{tabular}{|p{\textwidth}|}
        \hline
        \textbf{Question:} According to the 2000 United States census, what was the 2000 population of the birth city of the only 21st-century mayor of Austin, Texas who also served as mayor in the 1990s? Round your answer to the nearest thousand. \\
        \hline
        \textbf{Thought:} I need to determine the birth city of the only 21st-century mayor of Austin, Texas who also served as mayor in the 1990s and then find its population in the 2000 United States census. First, I will identify the mayor who meets this criterion. \\
        \textbf{Action:} search[Austin, Texas mayors; Who was the 21st-century mayor of Austin, Texas who also served as mayor in the 1990s?] \\
        \textbf{Observation:} (Result 1) Mayor of Austin -  Kirk Watson took office as mayor on January 6, 2023, for a second term, having served as mayor from 1997 to 2001. \\
        (Result 2) 2024 Austin mayoral election - Kirk Watson \\
        (Result 3) Austin, Texas - Kirk Watson \\
        \hline
        \textbf{Thought:} The only 21st-century mayor of Austin, Texas who also served as mayor in the 1990s is Kirk Watson. I now need to find his birth city. \\
        \textbf{Action:} search[Kirk Watson; Where was Kirk Watson born?] \\
        \textbf{Observation:} (Result 1) Kirk Watson - Oklahoma City, Oklahoma, U.S. \\
        (Result 2) Austin, Texas - Kirk Watson was born in 1958 in Oak Park, Illinois \\
        \hline
        \textbf{Thought:} The information provided contains conflicting birth cities for Kirk Watson. To confirm his birth city, I will perform another search. \\
        \textbf{Action:} search[Kirk Watson (American politician); What is the confirmed birth city of Kirk Watson?] \\
        \textbf{Observation:} (Result 1) Kirk Watson - Oklahoma City \\
        \hline
        \textbf{Thought:} Based on the confirmed information, Kirk Watson's birth city is Oklahoma City. Now, I need to find the population of Oklahoma City in the 2000 United States census. \\
        \textbf{Action:} search[Oklahoma City, Oklahoma; What was the population of Oklahoma City in the 2000 United States census?] \\
        \textbf{Observation:} (Result 1) Oklahoma City - 506,132 \\
        \hline
        \textbf{Thought:} The population of Oklahoma City in the 2000 United States census was 506,132. To answer the question, I need to round this number to the nearest thousand. \\
        \textbf{Action:} finish[506,000] \\
        \hline
    \end{tabular}
    \caption{An example of~\noteswriting~on the Frames dataset using LLama-3.1-70B.}
    \label{tab:example_frames_llama}
\end{table*}

\begin{table*}[htbp!]
    \centering
    \small
    \begin{tabular}{|p{\textwidth}|}
    \hline
    \textbf{Question:} Diago Costa played for which club when he was awarded the first FIFA World Cup Goal based on a VAR Decision? \\
    \hline
    \textbf{Thought:} I need to determine which club Diago Costa was playing for when he was awarded the first FIFA World Cup Goal based on a VAR Decision. First, I will gather information on the first FIFA World Cup Goal awarded based on a VAR Decision. \\
    \textbf{Action:} search[FIFA World Cup Goal; What was the first FIFA World Cup Goal awarded based on a VAR Decision, and who scored it?] \\
    \textbf{Observation:} (Result 1) 2022 FIFA World Cup - Ao Tanaka \\
    \hline
    \textbf{Thought:} It seems like the search result did not provide the relevant information about the first FIFA World Cup Goal awarded based on a VAR Decision. \\
    \textbf{Action:} search[FIFA World Cup VAR Decision; What was the first FIFA World Cup Goal awarded based on a VAR Decision, and who scored it?] \\
    \textbf{Observation:} (Result 1) 2018 FIFA World Cup - Diego Costa's first goal against Portugal became the first World Cup goal based on a VAR decision. \\
    (Result 2) 2022 FIFA World Cup - The first FIFA World Cup Goal awarded based on a VAR Decision was scored by Ao Tanaka of Japan against Spain. \\
    (Result 3) 2019 FIFA Women's World Cup - The first FIFA World Cup goal awarded based on a VAR decision was scored by French midfielder Antoine Griezmann in the 58th minute of a group stage match against Australia on June 16, 2018. \\
    (Result 4) Video assistant referee - The first VAR decision at the World Cup came on 16 June 2018 in a group stage match between France and Australia, where referee Andres Cunha awarded a penalty to France after consulting with the VAR. The resulting penalty kick was successfully converted by Antoine Griezmann of France, but the goal was initially disallowed for offside. However, the VAR team correctly advised the referee that the offside infraction was committed by an Australian player, not a French player, and the goal was awarded. \\
    \hline
    \textbf{Thought:} It seems like the search results provided conflicting information about the first FIFA World Cup Goal awarded based on a VAR Decision. However, according to Result 1 of the second search, Diego Costa's first goal against Portugal became the first World Cup goal based on a VAR decision. \\
    \hline
    \textbf{Action:} search[Diego Costa; Which club did Diego Costa play for in the 2018 FIFA World Cup?] \\
    \textbf{Observation:} (Result 1) Diego Costa - Atl\'etico Madrid \\
    \hline
    \textbf{Thought:} Diego Costa played for Atl\'etico Madrid when he was awarded the first FIFA World Cup Goal based on a VAR Decision. \\
    \hline
    \textbf{Action:} finish[Atl\'etico Madrid] \\
    \textbf{Observation:} Atl\'etico Madrid \\
    \hline
    \end{tabular}
    \caption{An Example of~\noteswriting~on Frames dataset using GPT-4o-mini.}
    \label{tab:example_frames}
\end{table*}

\renewcommand{\arraystretch}{1.5}
\begin{table*}[htbp!]
\centering
\small
\begin{tabular}{|p{\textwidth}|}
\hline
\textbf{Question:} If my future wife has the same first name as the 15th first lady of the United States' mother and her surname is the same as the second assassinated president's mother's maiden name, what is my future wife's name? \\
\hline
\textbf{Thought:} I need to determine the first name of the 15th first lady of the United States and the surname of the second assassinated president's mother's maiden name. Let's start by identifying the 15th first lady of the United States and her mother's name. \\
\textbf{Action:} search[15th first lady of the United States; Who is the 15th first lady and what is her mother's name?] \\
\textbf{Observation:} No relevant information, try a different search term. \\
\hline
\textbf{Thought:} I need to rephrase my search to find the correct information about the 15th first lady of the United States and her mother's name. I will search for the name of the 15th first lady directly. \\
\textbf{Action:} search[15th first lady of the United States; Who is the 15th first lady?] \\
\textbf{Observation:} No relevant information, try a different search term. \\
\hline
\textbf{Thought:} It seems that I'm having trouble finding information about the 15th first lady of the United States. I will now search for a list of all first ladies to identify her name and then find her mother's name. \\
\textbf{Action:} search[List of first ladies of the United States; What is the name of the 15th first lady?] \\
\textbf{Observation:} (Result 1) List of first ladies of the United States - The 15th first lady was Harriet Lane, who served from March 4, 1857, to March 4, 1861. \\
\hline
\textbf{Thought:} I have identified the 15th first lady of the United States as Harriet Lane. Now, I need to find information about her mother's name. \\
\textbf{Action:} search[Harriet Lane; What is the name of Harriet Lane's mother?] \\
\textbf{Observation:} (Result 1) Harriet Lane - Harriet Lane's mother was Jane Ann Buchanan Lane. \\
\hline
\textbf{Thought:} I have found that Harriet Lane's mother was named Jane Ann Buchanan Lane. Now, I need to identify the second assassinated president of the United States and find his mother's maiden name. \\
\textbf{Action:} search[second assassinated president of the United States; Who is the second assassinated president?] \\
\textbf{Observation:} (Result 1) List of United States presidential assassination attempts and plots - James A. Garfield was the second U.S. president to be assassinated, occurring on July 2, 1881, less than four months after he took office. \\
(Result 2) List of presidents of the United States by time in office - The second assassinated president is James A. Garfield. \\
\hline
\textbf{Thought:} I need to rephrase my search to find the correct information about James A. Garfield's mother's maiden name. I will search for information about James A. Garfield and his family. \\
\textbf{Action:} search[James A. Garfield's mother; What was the maiden name of James A. Garfield's mother?] \\
\textbf{Observation:} No relevant information, try a different search term. \\
\hline
\textbf{Thought:} It seems I'm having difficulty finding the maiden name of James A. Garfield's mother. I will search for biographies of James A. Garfield to find more details about his family. \\
\textbf{Action:} search[James A. Garfield biography; What is known about James A. Garfield's family?] \\
\textbf{Observation:} (Result 1) James A. Garfield - James Abram Garfield was born on November 19, 1831, in a log cabin in Moreland Hills, Ohio. He was the youngest of five children. Garfield married Lucretia Rudolph in 1858, and they had seven children, five of whom survived infancy. \\
\hline
\textbf{Thought:} I still need to find the maiden name of James A. Garfield's mother. I will focus my search specifically on his mother's name. \\
\textbf{Action:} search[James A. Garfield's family; What is the name of James A. Garfield's mother?] \\
\textbf{Observation:} (Result 1) James A. Garfield - James A. Garfield's mother was named Eliza. \\
\hline
\textbf{Thought:} I am struggling to find the maiden name of James A. Garfield's mother. I will search for historical records or references that detail Eliza Garfield's family background. \\
\textbf{Action:} search[Eliza Garfield biography; What is known about Eliza Garfield's family and maiden name?] \\
\textbf{Observation:} No relevant information, try a different search term. \\
\hline
\textbf{Thought:} It seems I am having persistent difficulties finding the maiden name of James A. Garfield's mother, Eliza. I will try searching for historical documentation or genealogical resources that might provide her maiden name. \\
\textbf{Action:} search[Eliza Garfield maiden name; What is the maiden name of James A. Garfield's mother?] \\
\textbf{Observation:} (Result 1) Abbie C. B. Robinson - The maiden name of President James A. Garfield's mother is Eliza Ballou. \\
\hline
\textbf{Thought:} I have found that the maiden name of James A. Garfield's mother is Eliza Ballou. Now, I have the necessary information to form the name of my future wife. Her first name is Jane (from Jane Ann Buchanan Lane) and her last name is Ballou (from Eliza Ballou). \\
\textbf{Action:} finish[Jane Ballou] \\
\hline
\end{tabular}
\caption{An example of~\noteswriting~on the Frames dataset using GPT-4o-mini.}
\label{tab:example_frames_gpt}
\end{table*}

\renewcommand{\arraystretch}{1.5}
\begin{table*}[htbp!]
    \centering
    \small
    \begin{tabular}{|p{\textwidth}|}
\hline
\textbf{Question:} What 2016 film stars actor Xavier Samuel and an actress who portrayed Vasilisa Dragomir in the film "Vampire Academy"? \\ \hline
\textbf{Thought:} I need to search for the film that stars Xavier Samuel, then find out who played Vasilisa Dragomir in "Vampire Academy" to identify the actress. Finally, I will check if both appeared in the same film in 2016. \\ \hline

\textbf{Action:} search[Xavier Samuel filmography] \\ \hline

\textbf{Observation:} Wikipedia Title: Xavier Samuel \\
Xavier Samuel (born 10 December 1983) is an Australian film and theatre actor. He has appeared in leading roles in the feature films Adore, September, Further We Search, Newcastle, The Loved Ones, Frankenstein, A Few Best Men, and played Riley Biers... \\
Wikipedia Title: Mohanlal filmography \\
Mohanlal is an Indian actor, producer, distributor, host, and playback singer who has starred in mainstream blockbuster and art-house films. He has made over 355 appearances in feature films during the span of his career, primarily in Malayalam... \\
Wikipedia Title: Benedict Samuel \\
Benedict Samuel is an Australian actor, writer, and director best known for playing Jervis Tetch / Mad Hatter in the Fox crime series Gotham and as Owen in The Walking... \\
Wikipedia Title: Gérard Depardieu \\
Gérard Xavier Marcel Depardieu is a multinational French actor, known to be one of the most prolific in film history. An icon of French cinema, considered a world star in... \\
Wikipedia Title: Jagadish filmography \\
P. V. Jagadish Kumar, known as Jagadish, is an Indian actor, screenwriter, and television presenter. He has starred in over 400 Malayalam... \\ \hline

\textbf{Thought:} The search still did not yield the specific filmography of Xavier Samuel. I will try searching for his name along with the year 2016 to see if I can find a film he starred in that year. \\ \hline

\textbf{Action:} search[Xavier Samuel 2016 film] \\ \hline

\textbf{Observation:} Wikipedia Title: Xavier Samuel \\
Xavier Samuel (born 10 December 1983) is an Australian film and theatre actor. He has appeared in leading roles in the feature films Adore, September, Further We Search, Newcastle, The Loved Ones, Frankenstein, A Few Best Men, and played Riley Biers... \\
Wikipedia Title: Mr. Church \\
Mr. Church is a 2016 American drama film directed by Bruce Beresford and written by Susan McMartin. The film stars Eddie Murphy as the title character with Britt Robertson, Xavier Samuel, Lucy Fry, Christian Madsen and Natascha McElhone also... \\
Wikipedia Title: Benedict Samuel \\
Benedict Samuel is an Australian actor, writer, and director best known for playing Jervis Tetch / Mad Hatter in the Fox crime series Gotham and as Owen in The Walking... \\
Wikipedia Title: Samuel Hunt (actor) \\
Samuel Caleb Hunt (born October 10, 1986) is an American actor best known for his reoccurring role as Greg ``Mouse'' Gerwitz in the television series Chicago Fire and Chicago P.D., playing Xavier Rosen in the third season of Empire, and for starring... \\
Wikipedia Title: Spin Out (film) \\
Spin Out is a 2016 Australian romantic comedy film directed by Tim Ferguson and Marc Gracie and starring Xavier Samuel and Morgan... \\ \hline
\end{tabular}
    \caption{An example of ReAcT on the HotpotQA dataset using GPT-4o-mini. It starts by repeatedly searching similar query.}
    \label{tab:example_hotpotqa_react_gpt}
\end{table*}

\renewcommand{\arraystretch}{1.5}
\begin{table*}[htbp!]
    \centering
    \small
    \begin{tabular}{|p{\textwidth}|}
\hline
\textbf{Thought:} I have found that Xavier Samuel stars in the film ``Spin Out'' in 2016. Now I need to find out who portrayed Vasilisa Dragomir in ``Vampire Academy.'' \\ \hline

\textbf{Action:} search[Vasilisa Dragomir Vampire Academy actress] \\ \hline

\textbf{Observation:} Wikipedia Title: Vampire Academy (film) \\
Vampire Academy is a 2014 American fantasy comedy horror film directed by Mark Waters from a screenplay by Daniel Waters, based on the first book of the novel series of the same name by Richelle Mead. The film stars an ensemble cast led by Zoey... \\
Wikipedia Title: Bloodlines (book series) \\
Bloodlines is a book series by Richelle Mead. It is a spin-off of Mead's Vampire Academy series. The books follow alchemist Sydney Sage. \\
After the incident with Rose and Dimitri, Sydney is being closely watched by the Alchemists (humans who help... \\
Wikipedia Title: List of fictional princesses \\
This is a list of fictional princesses that have appeared in various works of fiction. This list is organized by medium and limited to well-referenced, notable examples of fictional... \\ \hline

\textbf{Thought:} The search did not provide the specific actress who portrayed Vasilisa Dragomir. I will select the ``Vampire Academy'' film page to find the cast details. \\ \hline

\textbf{Action:} select[Vampire Academy (film)] \\ \hline

\textbf{Observation:} 2014 fantasy comedy horror film by Mark Waters \\
\begin{tabular}[t]{l}
| Vampire Academy | | \\
| Theatrical release poster | | \\
| Directed by | Mark Waters | \\
| Screenplay by | Daniel Waters | \\
| Based on | \textit{Vampire Academy} by Richelle Mead | \\
| Produced by | Don Murphy, Susan Montford, Michael Preger, Deepak Nayar | \\
| Cinematography | Tony Pierce-Roberts | \\
| Edited by | Chris Gill | \\
| Music by | Rolfe Kent | \\
| Production companies | Kintop Pictures, Reliance Entertainment, Angry Films, Preger Entertainment | \\
| Distributed by | The Weinstein Company | \\
| Release dates | February 7, 2014 (US), July 14, 2014 (UK straight-to-DVD) | \\
\end{tabular} \\
\textit{Vampire Academy} is a 2014 American fantasy comedy horror film directed by Mark Waters from a screenplay by Daniel Waters, based on the first book of the novel series by Richelle Mead. The film stars an ensemble cast led by Zoey Deutch as Rose Hathaway, a Dhampir, and Lucy Fry as Lissa Dragomir, her Moroi best friend, and follows their return to the St Vladimir's Academy boarding school, after being on the run for one year. \\ \hline

\textbf{Thought:} I found that Lucy Fry portrays Vasilisa Dragomir in ``Vampire Academy.'' Now, I will check if both Xavier Samuel and Lucy Fry starred in the same film in 2016. Since Xavier Samuel is in ``Spin Out,'' I need to find out if Lucy Fry is also in this film. \\ \hline

\textbf{Action:} search[Spin Out film cast] \\ \hline
\end{tabular}
    \caption{Continued example of ReAcT on the HotpotQA dataset using GPT-4o-mini. It focuses on "Spin Out" but misses the potential answer "Mr. Church".}
    \label{tab:example_hotpotqa_react_gpt_1}
\end{table*}
\renewcommand{\arraystretch}{1.5}
\begin{table*}[htbp!]
    \centering
    \small
    \begin{tabular}{|p{\textwidth}|}
\hline
\textbf{Observation:} Wikipedia Title: Spin Out (film) \\
Spin Out is a 2016 Australian romantic comedy film directed by Tim Ferguson and Marc Gracie and starring Xavier Samuel and Morgan... \\
Wikipedia Title: Spinning Out \\
Spinning Out is an American drama television series, created by Samantha Stratton, that premiered on Netflix on January 1, 2020. In February 2020, the series was canceled after one season. \\
The series stars Kaya Scodelario as Kat Baker, a young ice... \\
Wikipedia Title: Spin (2021 film) \\
Spin is a 2021 American comedy film released as a Disney Channel Original Movie (DCOM). It is directed by Manjari Makijany from a screenplay by Carley Steiner and Josh A. Cagan. Produced by Disney Channel Original Productions, the film stars... \\
Wikipedia Title: Spinning Man \\
Spinning Man is a 2018 thriller film directed by Simon Kaijser based on a novel by George Harrar and adapted to screenplay by Matthew Aldrich. It stars Guy Pearce, Pierce Brosnan, Minnie Driver, Alexandra Shipp, Clark Gregg and Odeya Rush. The film... \\
Wikipedia Title: Dog Man (film) \\
Dog Man is a 2025 American animated superhero comedy film based on Dav Pilkey's children's graphic novel series Dog Man, produced by DreamWorks Animation and Scholastic Entertainment, and distributed by Universal Pictures. It is a spin-off and story... \\ \hline

\textbf{Thought:} The search did not provide the cast details for ``Spin Out.'' I will select the ``Spin Out (film)'' page to find the cast. \\ 
\textbf{Action:} select[Spin Out (film)] \\ \hline

\textbf{Observation:} 2016 Australian film \\
\begin{tabular}[t]{l}
| Spin Out | | \\
| --- | --- | \\
| Theatrical film poster | | \\
| Directed by | Tim Ferguson, Marc Gracie | \\
| Written by | Edwina Exton, Tim Ferguson | \\
| Produced by | Marc Gracie, David Redman | \\
| Starring | Xavier Samuel, Morgan Griffin, Lincoln Lewis | \\
| Cinematography | Justin Brickle | \\
| Edited by | Ken Sallows | \\
| Music by | Jon Hume | \\
\end{tabular} \\
\textit{Spin Out} is a 2016 Australian romantic comedy film directed by Tim Ferguson and Marc Gracie and starring Xavier Samuel and Morgan Griffin. \\ \hline

\textbf{Thought:} The search results confirm that ``Spin Out'' stars Xavier Samuel, but there is no mention of Lucy Fry in the cast list. Thus, they did not appear in the same film. I will now summarize the findings to answer the question. \\ \hline

\textbf{Action:} finish[There is no common film between Xavier Samuel and Lucy Fry in 2016.] \\ \hline

\textbf{Observation:} There is no common film between Xavier Samuel and Lucy Fry in 2016. \\ \hline
\end{tabular}
    \caption{Continued example of ReAcT on the HotpotQA dataset using GPT-4o-mini. It fails by focusing exclusively on Xavier Samuel's film "Spin Out" while completely overlooking "Mr. Church," despite it being mentioned in the search results alongside both actors' names. This prevents it from from discovering that "Mr. Church" is the 2016 film starring both Xavier Samuel and Lucy Fry (who played Vasilisa Dragomir in "Vampire Academy").}
    \label{tab:example_hotpotqa_react_gpt_2}
\end{table*}

\renewcommand{\arraystretch}{1.5}
\begin{table*}[htbp!]
    \centering
    \small
    \begin{tabular}{|p{\textwidth}|}
        \hline
        \textbf{Question:} What 2016 film stars actor Xavier Samuel and an actress who portrayed Vasilisa Dragomir in the film "Vampire Academy"? \\
        \hline
        \textbf{Thought:} I need to identify the actress who portrayed Vasilisa Dragomir in "Vampire Academy" and then find a 2016 film that stars Xavier Samuel alongside her. First, I will search for the cast of "Vampire Academy" to determine who played Vasilisa Dragomir. \\
        \textbf{Action:} search[Vampire Academy cast; Who portrayed Vasilisa Dragomir in the film "Vampire Academy"?] \\
        \textbf{Observation:} (Result 1) Vampire Academy (film) - Lucy Fry portrayed Vasilisa "Lissa" Dragomir in the film "Vampire Academy". \\
        \hline
        \textbf{Thought:} Lucy Fry portrayed Vasilisa Dragomir in "Vampire Academy." Now, I need to find a 2016 film that stars both Xavier Samuel and Lucy Fry. \\
        \textbf{Action:} search[Xavier Samuel; What films did Xavier Samuel star in during 2016?] \\
        \textbf{Observation:} (Result 1) Xavier Samuel - In 2016, Xavier Samuel played the titular character in *The Death and Life of Otto Bloom*, and performed alongside Kate Beckinsale, Chloe Sevigny, Stephen Fry, and James Fleet in the critically-acclaimed film *Love and Friendship*, adapted from Jane Austen novel *Lady Susan*. \\
        \hline
        \textbf{Thought:} I found two films that Xavier Samuel starred in during 2016: *The Death and Life of Otto Bloom* and *Love and Friendship*. Now, I need to check if Lucy Fry appeared in either of these films. \\
        \textbf{Action:} search[Lucy Fry; What films did Lucy Fry star in during 2016?] \\
        \textbf{Observation:} (Result 1) Lucy Fry - In 2016, Lucy Fry starred in the films *Mr. Church* as Poppy and *The Darkness* as Stephanie Taylor. \\
        \hline
        \textbf{Thought:} Lucy Fry starred in *Mr. Church* and *The Darkness* in 2016. I need to check if either of these films also features Xavier Samuel. \\
        \textbf{Action:} search[Mr. Church; Did Xavier Samuel star in Mr. Church?] \\
        \textbf{Observation:} (Result 1) Mr. Church - Xavier Samuel is listed as one of the stars in the film *Mr. Church*. \\
        \hline
        \textbf{Thought:} I have confirmed that both Xavier Samuel and Lucy Fry starred in the film *Mr. Church* in 2016. \\
        \textbf{Action:} finish[*Mr. Church*] \\
        \textbf{Observation:} *Mr. Church* \\
        \hline
    \end{tabular}
    \caption{The previouis example with~\noteswriting~on the HotpotQA dataset using GPT-4o-mini.}
    \label{tab:example_hotpotqa_llama}
\end{table*}

\begin{table*}[htbp!]
    \centering
    \small
    \begin{tabular}{|p{\textwidth}|}
    \hline
    \textbf{Question:} As of August 1, 2024, what is the largest city of the 9th largest country by land area in Europe? \\
    \hline
    \textbf{Thought:} I need to determine the 9th largest country by land area in Europe and then find its largest city. First, I will identify the 9th largest country by land area in Europe. \\
    \textbf{Action:} search[Largest countries in Europe by land area; Which is the 9th largest country by land area in Europe?] \\
    \textbf{Observation:} (Result 1) List of countries and dependencies by area - The 9th largest country by land area in Europe is Kazakhstan (2,699,700 km\textsuperscript{2}). \\
    (Result 2) List of European countries by area - Poland \\
    \hline
    \textbf{Thought:} \textbf{It seems like there are multiple conflicting sources for the 9th largest country by land area in Europe, with one source saying it's Kazakhstan, but Kazakhstan is not in Europe, and another source saying it's Poland but without providing more information. In this case, I'll try to find the correct 9th largest country by land area in Europe}. \\
    \textbf{Action:} search[Largest countries in Europe by land area; What are the top 10 largest countries by land area in Europe?] \\
    \textbf{Observation:} (Result 1) List of countries and dependencies by area -  \\
    1. Russia - 16,376,870 km\textsuperscript{2} \\
    2. Ukraine - 579,330 km\textsuperscript{2} \\
    3. France - 640,427 km\textsuperscript{2} \\
    4. Spain - 498,980 km\textsuperscript{2} \\
    5. Sweden - 407,284 km\textsuperscript{2} \\
    6. Norway - 366,704 km\textsuperscript{2} \\
    7. Germany - 349,390 km\textsuperscript{2} \\
    8. Finland - 303,815 km\textsuperscript{2} \\
    9. Poland - 304,255 km\textsuperscript{2} \\
    10. Italy - 295,717 km\textsuperscript{2} \\
    \hline
    \textbf{Thought:} The 9th largest country by land area in Europe is indeed Poland, with a land area of 312,679 km\textsuperscript{2} or 304,255 km\textsuperscript{2}. Now, I need to determine the largest city of Poland. \\
    \textbf{Action:} search[Poland; What is the largest city in Poland as of August 1, 2024?] \\
    \textbf{Observation:} (Result 1) Poland - Warsaw \\
    (Result 2) Invasion of Poland - The largest city in Poland as of August 1, 2024, is Warsaw (Polish: Warszawa), which is also the capital of Poland. \\
    \hline
    \textbf{Thought:} The largest city of the 9th largest country by land area in Europe as of August 1, 2024, is Warsaw. \\
    \textbf{Action:} finish[The largest city of the 9th largest country by land area in Europe as of August 1, 2024, is Warsaw.] \\
    \hline
    \end{tabular}
    \caption{An example from the FRAMES dataset where note taking leads to conflicting information (in bold). To clarify the confusion the agent re-tries the search to find the correct information.}
    \label{tab:conflict_example_frames}
\end{table*}

\renewcommand{\arraystretch}{1.5}
\begin{table*}[htbp!]
    \centering
    \small
    \begin{tabular}{|p{\textwidth}|}
        \hline
        \textbf{Question:} A general motors vehicle is named after the largest ward in the country of Monaco.  How many people had walked on the moon as of the first model year of the vehicle? Note: the model year is not the same as the year the model was first produced. \\
        \hline
        \textbf{Thought:} I need to determine which General Motors vehicle is named after the largest ward in Monaco and find out the first model year of that vehicle. After that, I will find out how many people had walked on the moon by that year. Let's start by searching for the General Motors vehicle named after the largest ward in Monaco. \\
        \textbf{Action:} search[General Motors vehicle named after the largest ward in Monaco; What is the largest ward in Monaco and which GM vehicle is named after it?] \\
        \textbf{Observation:} No relevant information, try a different search term. \\
        \hline
        \textbf{Thought:} I need to find the largest ward in Monaco first and then see if there is a General Motors vehicle named after it. I will search for the largest ward in Monaco. \\
        \textbf{Action:} search[largest ward in Monaco; What is the largest ward in Monaco?] \\
        \textbf{Observation:} (Result 1) Monaco - The largest quarter in Monaco is Monte Carlo. \\
        \hline
        \textbf{Thought:} The largest ward in Monaco is Monte Carlo. Now, I need to search for any General Motors vehicle named after Monte Carlo and find out its first model year. \\
        \textbf{Action:} search[Monte Carlo; What is the General Motors vehicle named after Monte Carlo and what is its first model year?] \\
        \textbf{Observation:} (Result 1) Chevrolet Monte Carlo - The {Chevrolet Monte Carlo} is the General Motors vehicle named after Monte Carlo, and it was introduced for the 1970 model year. \\
        \hline
        \textbf{Thought:} The Chevrolet Monte Carlo is named after the largest ward in Monaco and was introduced for the 1970 model year. Now, I need to find out how many people had walked on the moon by the end of 1970. \\
        \textbf{Action:} search[moon landings; How many people had walked on the moon by the end of 1970?] \\
        \textbf{Observation:} (Result 1) Moon landing - \textbf{A total of twelve astronauts have landed on the Moon}. \\
        (Result 2) Apollo program - By the end of 1970, twelve people had walked on the Moon. \\
        \hline
        \textbf{Thought:} By the end of 1970, a total of twelve astronauts had walked on the Moon. Now I can provide the final answer regarding how many people had walked on the Moon as of the first model year of the Chevrolet Monte Carlo. \\
        \textbf{Action:} finish[12] \\
        \textbf{Observation:} 12 \\
        \hline
    \end{tabular}
    \caption{An example of hallucination by the note taking LLM in~\noteswriting~on the Frames dataset by GPT-4o-mini. The in-correct note taken is in bold. Instead the note writing LLM should have calculated the number of people walked before 1970 which is 4. }
    \label{tab:hallucinated_example_1}
\end{table*}

\renewcommand{\arraystretch}{1.5}
\begin{table*}[htbp!]
    \centering
    \small
    \begin{tabular}{|p{\textwidth}|}
        \hline
        \textbf{Question:} How old would the founder of the publishing company of the magazine that serialized the manga series Raw Hero have been the year the magazine ended publication? \\
        \hline
        \textbf{Thought:} I need to first identify the founder of the publishing company that serialized the manga series Raw Hero. Then I will determine the year the magazine ended publication and calculate the founder's age at that time. \\
        \textbf{Action:} search[Raw Hero; Which publishing company serialized the manga series Raw Hero?] \\
        \textbf{Observation:} (Result 1) Raw Hero - *Raw Hero* was serialized in Kodansha's *Evening*. \\
        \hline
        \textbf{Thought:} The manga series Raw Hero was serialized in Kodansha's *Evening* magazine. Now, I need to find out who the founder of Kodansha is. \\
        \textbf{Action:} search[Kodansha; Who is the founder of Kodansha?] \\
        \textbf{Observation:} (Result 1) Kodansha - Kodansha was founded by Seiji Noma in 1909. \\
        \hline
        \textbf{Thought:} Seiji Noma is the founder of Kodansha, which serialized *Raw Hero*. Next, I need to determine the year *Evening* magazine ended publication to calculate his age at that time. \\
        \textbf{Action:} search[Evening magazine; When did Evening magazine cease publication?] \\
        \textbf{Observation:} (Result 1) *Evening* (magazine) - The magazine ended publication on February 28, 2023. \\
        (Result 2) PM Magazine - *Evening Magazine* ceased publication on August 30, 1991. \\
        \hline
        \textbf{Thought:} The *Evening* magazine ceased publication on February 28, 2023. \textbf{I will use the founding year of Kodansha (1909) to calculate the age of Seiji Noma} at the time of the magazine's end in 2023. \\
        \textbf{Action:} finish[113] \\
        \textbf{Observation:} 113 \\
        \hline
    \end{tabular}
    \caption{An example of hallucination by the main LLM in~\noteswriting~on the Frames dataset by GPT-4o-mini. The LLM uses the founding year of the magazine as the birth year of the founder (highlighted in bold) instead of first identifying the birth year.}
    \label{tab:hallucinated_example_2}
\end{table*}

\begin{figure*}[h!]
    \tiny
    \begin{tcolorbox}[
        colback=gray!10, 
        colframe=black, 
        title=IRCoT and FLARE fewshot prompt
    ]
Task: Given a question and relevant documents, generate a step-by-step reasoning process to solve it.\\

Instructions:\\
- Enclose each reasoning step within <s> </s> tags.\\
- Reason in short sentences as the sentences will be used as queries to retrieve relevant documents for subsequent steps.\\
- If provided with a partial reasoning chain, complete it by generating the next set of steps.\\
- If you do not have up-to-date knowledge, try to think of a plan to solve the question.\\
- Provide the final answer inside <answer> </answer> tags.\\

DOCUMENTS:\\
Nobody Loves You (When You're Down and Out) - "Nobody Loves You (When You're Down and Out)" is a song written by John Lennon released on his 1974 album "Walls and Bridges". The song is included on the 1986 compilation "Menlove Ave.", the 1990 boxset "Lennon", the 1998 boxset "John Lennon Anthology", the 2005 two-disc compilation "", and the 2010 boxset "Gimme Some Truth".\\

John Lennon/Plastic Ono Band - John Lennon/Plastic Ono Band is the debut studio album by English rock musician John Lennon. It was released in 1970, after Lennon had issued three experimental albums with Yoko Ono and "Live Peace in Toronto 1969", a live performance in Toronto credited to the Plastic Ono Band. The album was recorded simultaneously with Ono's debut avant garde solo album, "Yoko Ono/Plastic Ono Band", at Ascot Sound Studios and Abbey Road Studios using the same musicians and production team and nearly identical cover artwork.\\

Walls and Bridges - Walls and Bridges is the fifth studio album by English musician John Lennon. It was issued by Apple Records on 26 September 1974 in the United States and on 4 October in the United Kingdom. Written, recorded and released during his 18-month separation from Yoko Ono, the album captured Lennon in the midst of his "Lost Weekend". "Walls and Bridges" was an American "Billboard" number-one album and featured two hit singles, "Whatever Gets You thru the Night" and "\#9 Dream". The first of these was Lennon's first number-one hit in the United States as a solo artist, and his only chart-topping single in either the US or Britain during his lifetime.\\

Question: Nobody Loves You was written by John Lennon and released on what album that was issued by Apple Records, and was written, recorded, and released during his 18 month separation from Yoko Ono?\\
Step-by-step reasoning:\\
<s>Identify album issued by Apple Records and recorded during John Lennon's 18-month separation from Yoko Ono.</s>\\
<s>The album "Walls and Bridges" was issued by Apple Records and recorded during this period.</s>\\
<s>Determine the album on which "Nobody Loves You (When You're Down and Out)" was released.</s>\\
<s>"Nobody Loves You (When You're Down and Out)" was released on "Walls and Bridges".</s>\\
<answer>Walls and Bridges</answer>\\

DOCUMENTS:\\
Route 13 (Laos) - Route 13 is the most important highway in the country of Laos. It begins at Boten in the North of Laos at the Chinese border. It connects the city of Vientiane to Luang Prabang in the north and roughly follows the line of the Mekong River down to the border with Cambodia. The road then continues at National Highway 7 in Cambodia. Also, this national highway leads past all three international airports in Laos: Vientiane Airport, Luang Prabang Airport, and Pakse Airport. Between Boten and Nateuy, Route 13 is at a length of 20 km as part of the Kunming-Bangkok Expressway, which is also part of the Asian Highway AH3. From Nateuy to Vientiane, Route 13 is part of AH12, and from Vientiane to the border of Cambodia, it is part of AH11.\\

National Route 13 (Vietnam) - National Route 13 is a highway in southern Vietnam stretching from the northeastern outskirts of Ho Chi Minh City, the commercial center of the country, towards the border to Cambodia. The highway starts around on the northern outskirts of Ho Chi Minh City, once the site of the military academy of the Army of the Republic of Vietnam, and travels north through the provinces.\\

Cambodia - Cambodia (Khmer: or Kampuchea), officially known as the Kingdom of Cambodia , is a country located in the southern portion of the Indochina Peninsula in Southeast Asia. It is 181,035 km² in area, bordered by Thailand to the northwest, Laos to the northeast, Vietnam to the east, and the Gulf of Thailand to the southwest.\\

Question: What is known as the Kingdom and has National Route 13 stretching towards its border?\\

Step-by-step reasoning:\\
<s>Cambodia is officially known as the Kingdom of Cambodia.</s>\\
<s>National Route 13 in Vietnam stretches towards the border of Cambodia.</s>\\
<s>Since Cambodia is referred to as the Kingdom and National Route 13 leads to its border, Cambodia is the correct answer.</s>\\
<answer>Cambodia</answer>\\

DOCUMENTS:\\
Vandenbrink GTO - The Vandenbrink GTO is a limited re-bodied version of the Ferrari 599 GTB Fiorano. It features an entirely new coachwork designed by Michiel van den Brink of Vandenbrink Design, inspired by the 1962 Ferrari 250 GTO as a tribute.\\

1969 World Sportscar Championship - The 1969 World Sportscar Championship was the 17th season of FIA World Sportscar racing, featuring Group 6 Prototype Sports Cars, Group 4 Sports Cars, and Group 3 Grand Touring Cars.\\

Ferrari 250 GTO - The Ferrari 250 GTO was a GT car produced by Ferrari from 1962 to 1964 for homologation into the FIA's Group 3 Grand Touring Car category. It was powered by Ferrari's Tipo 168/62 V12 engine.\\

Question: Which car, produced by Ferrari from 1962 to 1964 for homologation into the FIA's Group 3 Grand Touring Car category, inspired the Vandenbrink GTO?\\

Step-by-step reasoning:\\
<s>The Ferrari 250 GTO was produced by Ferrari from 1962 to 1964 for homologation into the FIA's Group 3 Grand Touring Car category.</s>\\
<s>The Vandenbrink GTO was inspired by the Ferrari 250 GTO.</s>\\
<s>Since the question asks for the car that inspired the Vandenbrink GTO, the correct answer is the Ferrari 250 GTO.</s>\\
<answer>Ferrari 250 GTO</answer>\\

DOCUMENTS:\\
Talula - "Talula" is a song by Tori Amos, released as the second single from her 1996 album "Boys For Pele". It reached \#22 on the UK Singles Chart and appears in the Jan de Bont film "Twister".\\

Nema aviona za Zagreb - "Nema aviona za Zagreb" ("There Is No Plane to Zagreb") is a 2012 Dutch film by Louis van Gasteren. The film is a retrospective of events in the director's life from 1964 to 1969, filmed by him in that period and reflected on from his vantage point over 40 years later at the age of 90. It is also Jan de Bont's first film credit as cinematographer since 1992's "Basic Instinct".\\

Jan de Bont - Jan de Bont (born 22 October 1943) is a Dutch cinematographer, director, and film producer. He is widely known for directing the 1994 action film "Speed," starring Keanu Reeves and Sandra Bullock, and the disaster film "Twister." As a director of photography, he has shot numerous blockbusters and genre films, including "Cujo, Flesh and Blood, Die Hard, The Hunt for Red October," and "Basic Instinct."\\

Question: Who was born first? Jan de Bont or Raoul Walsh?\\

Step-by-step reasoning:\\
<s>Jan de Bont was born on October 22, 1943.</s>\\
<s>Raoul Walsh was born on March 11, 1887.</s>\\
<s>Since 1887 is earlier than 1943, Raoul Walsh was born first.</s>\\
<answer>Raoul Walsh</answer>
    \end{tcolorbox}
    \caption{Fewshot prompt used for the evaluation of IRCoT and FLARE methods.}
    \label{fig:ircot_prompt}
\end{figure*}

\begin{figure*}[h!]
\small
\begin{tcolorbox}[colback=gray!10, colframe=black, title=Prompt for Notes Writing]
Extract relevant information which is not previously extracted from the Wikipedia page provided in markdown format relevant to the given query. You will be provided with the Wikipedia page, query, and the previously extracted content. Do not miss any information. Do not add irrelevant information or anything outside of the provided sources.

Provide the answer in the format: \texttt{<YES/NO>\#<Relevant context>}.

Here are the rules:
\begin{itemize} [noitemsep, nolistsep]
    \item If you don't know how to answer the query - start your answer with \texttt{NO\#}
    \item If the text is not related to the query - start your answer with \texttt{NO\#}
    \item If the content is already extracted - start your answer with \texttt{NO\#}
    \item If you can extract relevant information - start your answer with \texttt{YES\#}
\end{itemize}

Example answers:
\begin{itemize} [noitemsep, nolistsep]
    \item \texttt{YES\#Western philosophy originated in Ancient Greece in the 6th century BCE with the pre-Socratics.}
    \item \texttt{NO\#No relevant context.}
\end{itemize}

\textbf{Context:} \texttt{\{Context\}}

\textbf{Previous Context:} \texttt{\{PrevContext\}}

\textbf{Query:} \texttt{\{Query\}}

\end{tcolorbox}
\label{fig:notes_writing_prompt}
\caption{Notes writing prompt for extracting the relevant information.}
\end{figure*}

\begin{figure*}[h!]
\small
\begin{tcolorbox}[colback=gray!10, colframe=black, title=GPT-4 Judge Prompt]
\textbf{===Task===} \\
I need your help in evaluating an answer provided by an LLM against a ground truth answer. Your task is to determine if the ground truth answer is present in the LLM's response. Please analyze the provided data and make a decision.

\textbf{===Instructions===}
\begin{enumerate} [noitemsep, nolistsep]
    \item Carefully compare the "Predicted Answer" with the "Ground Truth Answer".
    \item Consider the substance of the answers – look for equivalent information or correct answers. Do not focus on exact wording unless the exact wording is crucial to the meaning.
    \item Your final decision should be based on whether the meaning and the vital facts of the "Ground Truth Answer" are present in the "Predicted Answer."
\end{enumerate}

\textbf{===Input Data===}
\begin{itemize} [noitemsep, nolistsep]
    \item \textbf{Question:} \texttt{<<question>>}
    \item \textbf{Predicted Answer:} \texttt{<<LLM\_response>>}
    \item \textbf{Ground Truth Answer:} \texttt{<<ground\_truth\_answer>>}
\end{itemize}

\textbf{===Output Format===} \\
Provide your final evaluation in the following format:

\textbf{Explanation:} (How you made the decision?) \\
\textbf{Decision:} ("TRUE" or "FALSE")

Please proceed with the evaluation.
\end{tcolorbox}
\caption{GPT-4 prompt for evaluating the correctness of predicted answer.}
\label{fig:judge_prompt}
\end{figure*}

\begin{figure*}[h!]
    \small
    \begin{tcolorbox}[
        colback=gray!10, 
        colframe=black, 
        title=Quality evaluation prompt
    ]

You are asked to evaluate the reasoning chain produced in response to a question, particularly focusing on how effectively tools were used throughout the process. The evaluation should be based on the following clearly defined criteria. For each criterion, provide a numerical rating on a scale from 0 to 5, where 5 represents excellent performance and 0 indicates poor or entirely absent performance.\\
\\
Criterion 1: Efficiency of the Steps Taken\\
Definition:\\
Evaluate the overall efficiency of each step in the reasoning chain, with specific focus on whether the tool calls and reasoning steps helped progress toward the final correct answer. Efficient steps reduce uncertainty, narrow the solution space, or directly contribute to solving the problem.\\
Rating Guide:\\
5 – Extremely efficient: Every step clearly advances the reasoning; no wasted effort.\\
4 – Highly efficient: Most steps are purposeful, with only minor inefficiencies.\\
3 – Moderately efficient: Some steps are valuable, others contribute little.\\
2 – Minimally efficient: Several steps are misdirected or low-impact.\\
1 – Poorly efficient: Most steps offer minimal or no progress toward the answer.\\
0 – Not efficient at all: Steps are irrelevant, aimless, or distracting.\\
\\
Criterion 2: Redundancy of Steps\\
Definition:\\
Assess the reasoning chain for unnecessary repetition or duplication of steps, including redundant tool calls or rephrasing of the same logic without new insight. A low-redundancy chain avoids rework and keeps the progression streamlined.\\
Rating Guide:\\
5 – No redundancy: Each step is unique and adds distinct value.\\
4 – Very low redundancy: Only minor repetition, quickly resolved.\\
3 – Moderate redundancy: Some ideas or tool uses are repeated without added benefit.\\
2 – Noticeable redundancy: Multiple steps repeat similar content or actions unnecessarily.\\
1 – High redundancy: Repetition significantly detracts from conciseness.\\
0 – Extremely redundant: Most of the chain rehashes prior reasoning with no new value.\\
\\
Criterion 3: Clarity and Coherence of the Reasoning Chain\\
Definition:\\
Examine how clearly and logically the reasoning chain progresses from the question to the final answer. This includes whether steps are easy to follow, logically connected, and free of ambiguity or excessive complexity.\\
Rating Guide:\\
5 – Exceptionally clear and coherent: The reasoning is logical, concise, and easy to follow.\\
4 – Mostly clear: The chain is understandable with minor clarity issues.\\
3 – Moderately clear: Some transitions or justifications are unclear or weak.\\
2 – Confusing in parts: Multiple unclear, inconsistent, or disjointed steps.\\
1 – Difficult to follow: Lacks logical flow or clear structure.\\
0 – Incomprehensible: The chain cannot be understood or followed logically.\\
\\
First provide your reasoning of your evaluation then structure your responses as a json with the keys "Criterion 1", "Criterion 2", "Criterion 3" and the values as the ratings you provided.\\
Chain: \{\}

    \end{tcolorbox}
    \caption{Prompt for quality evaluation of reasoning chain.}
    \label{fig:quality_prompt}
\end{figure*}